\documentclass[11pt,a4paper]{article}
\usepackage[hyperref]{acl2019}
\usepackage{times}
\usepackage{latexsym}

\usepackage[T2A, T1]{fontenc}
\usepackage[utf8]{inputenc}

\usepackage{url}
\usepackage{graphicx}
\usepackage{multirow}
\usepackage{amsthm}
\usepackage{amssymb}
\usepackage{amsmath}
\usepackage{amsfonts}
\usepackage{xcolor}
\usepackage{xspace}
\usepackage{mdwlist}
\usepackage{microtype}
\usepackage{tabularx}
\usepackage{booktabs}
\usepackage{subcaption}
\usepackage{bbold}
\usepackage[multiple]{footmisc}
\usepackage[ruled,vlined]{algorithm2e}
\usepackage{arydshln}
\usepackage{tablefootnote}

\DeclareMathOperator*{\argmin}{arg\,min}

\usepackage{url}
\aclfinalcopy
\newcommand{\en}{\textsc{en}\xspace}
\newcommand{\de}{\textsc{de}\xspace}
\newcommand{\ita}{\textsc{it}\xspace} % \it is already defined
\newcommand{\fr}{\textsc{fr}\xspace}
\newcommand{\hr}{\textsc{hr}\xspace}
\newcommand{\ru}{\textsc{ru}\xspace}
\newcommand{\fin}{\textsc{fi}\xspace} % \fi is already defined
\newcommand{\tr}{\textsc{tr}\xspace}
\newcommand{\cca}{\textsc{CCA}\xspace}
\newcommand{\proc}{\textsc{Proc}\xspace}
\newcommand{\procb}{\textsc{Proc-B}\xspace}
\newcommand{\dlv}{\textsc{DLV}\xspace}
\newcommand{\rcsls}{\textsc{RCSLS}\xspace}
\newcommand{\vecmap}{\textsc{VecMap}\xspace}
\newcommand{\muse}{\textsc{Muse}\xspace}
\newcommand{\icp}{\textsc{ICP}\xspace}
\newcommand{\gwa}{\textsc{GWA}\xspace}

\title{How to (Properly) Evaluate Cross-Lingual Word Embeddings: \\ On Strong Baselines, Comparative Analyses, and Some Misconceptions}

\author{Goran Glava\v{s}\textsuperscript{1}, Robert Litschko\textsuperscript{1}, Sebastian Ruder\textsuperscript{2,3}, and Ivan Vuli\'{c}\textsuperscript{4,5} \vspace{1em} \\
  \textsuperscript{1}Data and Web Science Group, University of Mannheim, Germany \\
  \textsuperscript{2}Insight Research Centre, National University of Ireland, Galway, Ireland \\
  \textsuperscript{3}Aylien Ltd., Dublin, Ireland \\
  \textsuperscript{4} Language Technology Lab, University of Cambridge, UK \hspace{2mm} \textsuperscript{5} PolyAI Ltd., UK\\
  {\tt \{goran, rlitschk\}@informatik.uni-mannheim.de}, \\ 
  {\tt sebastian@ruder.io, iv250@cam.ac.uk}
}

\date{}

\begin{document}
\maketitle
\begin{abstract}

Cross-lingual word embeddings (CLEs) enable multilingual modeling of meaning and facilitate cross-lingual transfer of NLP models. Despite their ubiquitous usage in downstream tasks, recent increasingly popular projection-based CLE models are almost exclusively evaluated on a single task only: bilingual lexicon induction (BLI). Even BLI evaluations vary greatly, hindering our ability to correctly interpret performance and properties of different CLE models. In this work, we make the first step towards a comprehensive evaluation of cross-lingual word embeddings. We thoroughly evaluate both supervised and unsupervised CLE models on a large number of language pairs in the BLI task and three downstream tasks, providing new insights concerning the ability of cutting-edge CLE models to support cross-lingual NLP. We empirically demonstrate that the performance of CLE models largely depends on the task at hand and that optimizing CLE models for BLI can result in deteriorated downstream performance. We indicate the most robust supervised and unsupervised CLE models and emphasize the need to reassess existing baselines, which still display competitive performance across the board. We hope that our work will catalyze further work on CLE evaluation and model analysis.

\end{abstract}

\section{Introduction and Motivation}
%%
% definition of the CLE and models
%%
Following the ubiquitous use of word embeddings in monolingual NLP tasks, many researchers have broadened their interest towards \textit{cross-lingual word embeddings} (CLEs). CLE models learn vectors of words in two or more languages and represent them in a \textit{shared cross-lingual word vector space}, where words with similar meanings obtain similar vectors, irrespective of their language. Owing to this property, CLEs hold promise to support cross-lingual NLP by enabling multilingual modeling of meaning and facilitating cross-lingual transfer for downstream NLP tasks and resource-poor languages. They serve as an invaluable source of (cross-lingual) knowledge in tasks such as bilingual lexicon induction \cite{mikolov2013exploiting,heyman2017bilingual}, document classification \cite{klementiev2012inducing}, information retrieval \cite{vulic2015sigir}, dependency parsing \cite{guo2015cross}, sequence labeling \cite{zhang2016ten,mayhew2017cheap}, and machine translation \cite{artetxe2018unsupervised,lample2018phrase}, among others.

%Cross-lingual word embeddings (CLEs) are vector spaces in which points correspond to words from two or more languages. In a (meaningful) cross-lingual word vector space, words with similar meaning, whether from the same language or different ones, should have similar vectors. 

Earlier work typically induces CLEs by leveraging bilingual supervision from multilingual corpora aligned at the level of sentences \cite[\textit{inter alia}]{klementiev2012inducing,zou2013bilingual,hermann2014multilingual,luong2015bilingual,gouws2015bilbowa} and documents \cite[\textit{inter alia}]{Sogaard:2015acl,vulic2016bilingual,levy2017strong}. A recent trend, however, are the so-called \textit{projection-based approaches}\footnote{In the literature the methods are sometimes referred to as mapping-based CLE approaches or offline approaches.} that post-hoc align pre-trained monolingual embeddings. Their popularity stems from their competitive performance coupled with a conceptually simple design, which requires only cheap bilingual supervision \cite{Ruder2018survey}. In particular, they demand word-level supervision from \textit{seed translation dictionaries} that span several thousand word pairs \cite{mikolov2013exploiting,faruqui2014improving,huang2015translation}, but it has also been shown that reliable projections can be bootstrapped from small dictionaries of 50--100 pairs \cite{vulic2016on,zhang2016ten}, identical strings and cognates \cite{hauer2017bootstrapping,smith2017offline,sogaard2018on}, and shared numerals \cite{artetxe2017learning}. 

In addition, recent work has leveraged topological similarities between monolingual vector spaces to introduce \textit{fully unsupervised} projection-based CLE approaches that do not require any bilingual supervision \cite{zhang2017adversarial,conneau2018word,artetxe2018robust,alvarez2018gromov}. Being conceptually attractive, such weakly supervised and unsupervised CLEs have taken the field by storm recently \cite{conneau2018word,Grave:2018arxiv,dou2018unsupervised,doval2018improving,hoshen2018nonadversarial,joulin2018loss,kim2018improving,chen2018unsupervised,mukherjee2018learning,nakashole2018norma,xu2018unsupervised,alaux2019unsupervised}. 

%% (IV, REMOVED, CITED BEFORE): alvarez2018gromov,artetxe2018robust,

% ruder2018discriminative,

%% criticism of evaluation
Regardless of their underlying modeling assumptions as well as data and supervision requirements, all CLE models are directly comparable as they produce the same ``end product'': a shared cross-lingual word vector space. %Therefore, they can support exactly the same groups of tasks.
Yet, a proper and comprehensive evaluation of recent CLE models is missing. Limited CLE evaluations impede comparative analyses and even lead to inadequate conclusions, as the models are typically overfitted to a single task.

Whereas early CLE models \cite{klementiev2012inducing,hermann2014multilingual} were evaluated in downstream tasks (primarily cross-lingual text classification), a large body of recent work is judged exclusively on the task of bilingual lexicon induction (BLI). This limits our understanding of CLE methodology as: \textbf{1)} BLI is an intrinsic task, and it has been shown that word translation performance and downstream performance on text classification and parsing tend to correlate poorly \cite{ammar2016massively}; \textbf{2)} We argue that BLI is not the main reason why we induce cross-lingual embedding spaces---rather, we seek to employ CLEs to tackle multilinguality and language transfer in downstream tasks. Therefore, it is not at all clear whether and to which extent BLI performance of (projection-based) CLE models correlates with various downstream tasks. In other words, previous research does not evaluate the true capacity of CLE models to support cross-lingual NLP.

Downstream evaluations aside, it is virtually impossible to directly compare all recently proposed mapping-based CLE models based on their published BLI results due to the lack of a common evaluation protocol: different papers consider different language pairs and employ different training and evaluation dictionaries. Furthermore, there is a surprising lack of testing BLI results for statistical significance. The mismatches in evaluation yield partial conclusions and inconsistencies: on the one hand, some unsupervised models \cite{artetxe2018robust,hoshen2018nonadversarial,alvarez2018gromov} are reported to outperform (or at least perform on par with) previous best-performing supervised CLE models \cite{artetxe2017learning,smith2017offline}. On the other hand, the most recent supervised approaches \cite{doval2018improving,joulin2018loss} report further performance gains, surpassing unsupervised models.

Supervised projection-based CLEs commonly require only small to moderately sized translation dictionaries (e.g. 1K--5K word translation pairs) and can also be bootstrapped from even smaller dictionaries. Such supervision data are easily obtainable for most language pairs.\footnote{One could argue that if a small word translation dictionary cannot be obtained for a pair of languages, one is probably dealing with truly under-resourced languages for which it would be difficult to obtain a large corpus required to train reliable monolingual embeddings in the first place. Furthermore, there are initiatives in typological linguistics research such as the ASJP database \cite{Wichmann:2018asjp}, which offers 40-item word lists denoting the same set of concepts in all the world's languages: \url{https://asjp.clld.org/}. Indirectly, such lists can offer the initial seed supervision.} Therefore, despite their attractive zero-supervision setup, we perceive unsupervised CLE models practically justified only if such models can, contrary to the intuition, indeed outperform supervised projection-based CLEs.      

\vspace{1.4mm}
\noindent \textbf{Contributions.} In this work, we provide a comprehensive comparative evaluation of a wide range of state-of-the-art---both supervised and unsupervised---projection-based CLE models. Our evaluation benchmark encompasses BLI and three cross-lingual downstream tasks of different nature: document classification (CLDC), information retrieval (CLIR), and natural language inference (XNLI). We unify evaluation protocols for all the representative models in our comparison, and conduct experiments on a large set of 28 language pairs that span diverse language types. 

In addition to providing a unified testbed for guiding CLE research, we primarily aim at answering the following two research questions: \textbf{1)} Is BLI performance a good predictor of downstream performance for projection-based CLE models? \textbf{2)} Can unsupervised CLE models indeed outperform their supervised counterparts? In many of our experiments, the simplest among the evaluated CLE methods outperform their more intricate competitors. We find that overfitting to BLI may severely hurt downstream performance, indicating that BLI evaluation should always be coupled with downstream evaluations in order to paint a more informative picture of CLE models' properties. 

%% IV REMOVED: (Too specific imho)
%We further inspect properties of cross-lingual embedding spaces induced by some of the models under evaluation and provide explanations for models' performance on downstream tasks.

%\input{2-RW_CLE.tex}
\begin{figure*}[t!]
    \centering
    \includegraphics[width=0.88\linewidth]{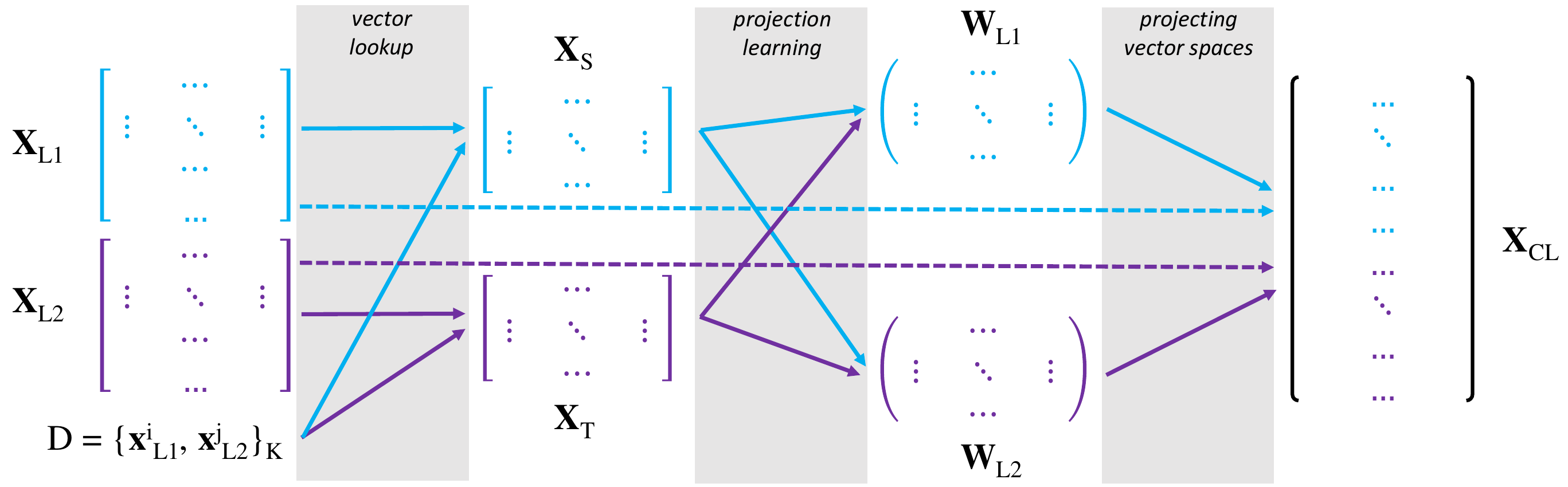}
    \vspace{-1.5mm}
    \caption{A general framework for post-hoc projection-based induction of cross-lingual word embeddings.}
    \label{fig:framework}
    \vspace{-3mm}
\end{figure*}

\section{Projection-Based Cross-Lingual Word Embeddings: Methodology}
In contrast to the more recent unsupervised models, CLE models typically require aligned words, sentences, or documents to learn a shared vector space. Supervised CLE models based on sentence- and document-aligned texts have been extensively studied and categorized according to the nature of required bilingual supervision in previous work \cite{vulic2016on,upadhyay2016acl,Ruder2018survey}. For a systematic overview of these earlier (resource-demanding) CLE models, we refer the reader to the survey papers. 
Current CLE research is almost exclusively focused on projection-based CLE models; they are therefore also the focal point of our comparative study.\footnote{Since these methods \textbf{a)} are not bound to any particular word embedding model (i.e., they are fully agnostic to how we obtain monolingual vectors) and \textbf{b)} they do not require any multilingual corpora, they lend themselves to a wider spectrum of languages than the alternatives \cite{Ruder2018survey}.}

\subsection{Projection-Based Framework}
%% IV REMOVED (this footnote is not very useful imho): \footnote{The figure depicts an induction of a bilingual embedding space, i.e., a cross-lingual embedding space between two languages.}
The goal is to learn a projection between independently trained monolingual word vector spaces. The mapping is sought using a seed bilingual lexicon, provided beforehand or extracted without supervision. A general post-hoc projection-based CLE induction framework is depicted in Figure \ref{fig:framework}. Let $\mathbf{X}_{L1}$ and $\mathbf{X}_{L2}$ be the monolingual embedding spaces of the two languages with respective vocabularies $V_{L1}$ and $V_{L2}$. All projection-based CLE models encompass the following steps: 
%\begin{enumerate*}
%% by comparing monolingual embedding spaces $\mathbf{X}_{L1}$ and $\mathbf{X}_{L2}$

\vspace{1.4mm}
\noindent \textbf{Step 1:} Construct the seed translation dictionary $D = \{(w^i_{L1}, w^j_{L2)}\}^{K}_{k = 1}$ containing $K$ word pairs. Supervised models simply use an external dictionary; unsupervised models induce $D$ automatically, usually assuming (approximate) isomorphism between monolingual spaces.

\vspace{1.4mm}    
\noindent \textbf{Step 2:} Create aligned monolingual subspaces (i.e., matrices) $\mathbf{X}_S = \{\mathbf{x}^i_{L1}\}^{K}_{k = 1}$ and $\mathbf{X}_T = \{\mathbf{x}^j_{L2}\}^{K}_{k = 1}$ by retrieving monolingual vectors of words from the translation dictionary: vectors of $\{w^i_{L1}\}^{K}_{k = 1}$ from $\mathbf{X}_{L1}$ and vectors of $\{w^j_{L2}\}^{K}_{k = 1}$ from $\mathbf{X}_{L2}$.
    
\vspace{1.4mm}
\noindent \textbf{Step 3:} Learn to project to the joint cross-lingual space using the word-aligned matrices $\mathbf{X}_S$ and $\mathbf{X}_T$. In the general case, we learn two projection matrices, $\mathbf{W}_{L1}$ and $\mathbf{W}_{L2}$, projecting respective vectors from $\mathbf{X}_{L1}$ and $\mathbf{X}_{L2}$ to the shared space $\mathbf{X}_{CL}$. Most models, however, directly project source language vectors from $\mathbf{X}_{L1}$ to the target language space $\textbf{X}_{L2}$ (i.e., $\mathbf{W}_{L2} = I$ and $\textbf{X}_{CL} = \textbf{X}_{L1}\textbf{W}_{L1} \cup \textbf{X}_{L2}$). 
%instead of projecting both $\mathbf{X}_{L1}$ and $\mathbf{X}_{L2}$ to a new space. 
%These models can trivially induce a multilingual space of $n$ languages through $n-1$ projections to the same pivot target language. 
%\end{enumerate*}

%% IV REMOVED (no need to cite the models yet again): \cite{zhang2017adversarial,conneau2018word,artetxe2018robust,alvarez2018gromov,hoshen2018nonadversarial} 

\subsection{Projection-Based CLE Models}

We evaluate a range of both supervised and unsupervised projection-based CLE models. While supervised models \cite{mikolov2013exploiting,smith2017offline,ruder2018discriminative,joulin2018loss} learn the projections using existing dictionaries, unsupervised models first induce seed dictionaries without bilingual data. We include unsupervised models with diverse dictionary induction strategies: adversarial learning \cite{conneau2018word}, similarity-based heuristics \cite{artetxe2018robust}, PCA \cite{hoshen2018nonadversarial}, and optimal transport \cite{alvarez2018gromov}. We next briefly describe the CLE models in our evaluation. %in more detail all of the CLE models evaluated in this study.

% discuss: supervised models, unsupervised models
%In terms of unsupervised approaches, aim to identify a good initialization: Via a GAN \cite{zhang2017adversarial,conneau2018word}, or heuristic \cite{artetxe2018robust} or an existing alignment method such as PCA \cite{hoshen2018nonadversarial} and optimal transport. 

%% (IV, something similar said before): Different to direct projection methods, learn only one projection from $\mathbf{X}_{L1}$ to $\mathbf{X}_{L2}$,

\vspace{1.4mm}
\noindent \textbf{Canonical Correlation Analysis (\cca).} \newcite{faruqui2014improving} use CCA to project $\mathbf{X}_{L1}$ and $\mathbf{X}_{L2}$ into a shared space $\mathbf{X}_{CL}$. \cca learns both $W_{L1}$ and $W_{L1}$, one for mapping each of the input spaces to the shared space. We evaluate \cca as a simple supervised baseline model that has mostly been neglected in recent BLI evaluations.

\vspace{1.4mm}
\noindent\textbf{Solving the Procrustes Problem.} In their seminal projection-based CLE work, \newcite{mikolov2013exploiting} cast the problem as learning the projection $\mathbf{W}_{L1}$ that minimizes the Euclidean distance between the projection of $\mathbf{X}_S$ and $\mathbf{X}_T$: 

\vspace{-1mm}
{\small
\begin{equation}
\label{eq:linmap_opt}
    \textbf{W}_{L1} = \argmin_{\mathbf{W}}\lVert \mathbf{X}_{L1} \mathbf{W} - \mathbf{X}_{L2} \rVert_2  
\end{equation}}%

\noindent They also report non-linear neural projections yielding worse BLI performance than a linear map $\mathbf{W}_{L1}$. \newcite{xing2015normalized} later achieves further BLI gains by constraining $\mathbf{W}_{L1}$ to an orthogonal matrix, ensuring that the topology of the original monolingual space is preserved. By imposing the orthogonality constraint, the optimization problem from Eq.~\eqref{eq:linmap_opt} becomes the Procrustes problem, with the following closed-form solution \cite{schonemann1966generalized}:

\vspace{-2mm}
{\small
\begin{align}
    \mathbf{W}_{L1} &= \mathbf{UV}^\top, \, \text{with} \notag \\
    \mathbf{U\Sigma V}^\top &= \mathit{SVD}(\mathbf{X}_{T} {\mathbf{X}_{S}}^\top).
    \label{eq:proc}
\end{align}}%
\begin{algorithm}[b]
{\footnotesize
 $\mathbf{X}_{L1}$, $\mathbf{X}_{L2} \leftarrow$ monolingual embeddings of $L1$ and $L2$ \\
 $D$ $\leftarrow$ initial word translation dictionary \\
 $n$ $\leftarrow$ number of bootstrapping iterations \\
 \For{each of $n$ iterations}{
   $\mathbf{X}_S \leftarrow$ \text{L1 vectors for left words in $D$} \\
   $\mathbf{X}_T \leftarrow$ \text{L2 vectors for right words in $D$} \\
   $\mathbf{W}_{L1} \leftarrow \argmin_{W}\lVert \mathbf{X}_S \mathbf{W} - \mathbf{X}_T \rVert_2$ \\
   $\mathbf{W}_{L2} \leftarrow \argmin_{W}\lVert \mathbf{X}_T \mathbf{W} - \mathbf{X}_S \rVert_2$ \\    
   \If{last iteration}{break}
   $\mathbf{X'}_{L1} \leftarrow \mathbf{X}_{L1} \mathbf{W}_{L1}$ \\
   $\mathbf{X'}_{L2} \leftarrow \mathbf{X}_{L2} \mathbf{W}_{L2}$ \\
   $D_{1\rightarrow 2} \leftarrow$ \text{most-similar}($V_{L1}$, $\mathbf{X'}_{L1}$,
   $\mathbf{X}_{L2}$) \\
   $D_{2\rightarrow 1} \leftarrow$ \text{most-similar}($V_{L2}$, $\mathbf{X'}_{L2}$, $\mathbf{X}_{L1}$) \\
   $D \leftarrow D \cup (D_{1\rightarrow 2} \cap D_{2\rightarrow 1})$}
   \text{return: }$\mathbf{W}_{L1}$ (\text{and/or} $\mathbf{W}_{L2}$)
 \caption{{\footnotesize{Bootstrapping Procrustes (\procb)}}}
}
\end{algorithm}
\noindent We use the linear map obtained as the solution to the Procrustes problem as the primary baseline in our evaluation (\proc). Furthermore, inspired by the bootstrapping \textit{self-learning} procedures that some unsupervised models employ to (iteratively) augment the induced lexicon $D$ \cite{artetxe2018robust,conneau2018word}, we propose a simple bootstrapping-based extension of the supervised \proc model (dubbed \procb). The intuition is that the bootstrapping approach can boost performance when smaller supervised lexicons are used. The procedure is summarized in Algorithm 1.

In each iteration, we use the translation lexicon $D$ to learn two unidirectional projections---$\mathbf{W}_{L1}$ that projects the embedding space $\mathbf{X}_{L1}$ of L1 to the embedding space $\mathbf{X}_{L2}$ of L2 and $\mathbf{W}_{L2}$ that, inversely, projects vectors from $\mathbf{X}_{L2}$ to $\mathbf{X}_{L1}$. We next create two sets of word translation pairs by comparing (1) $\mathbf{X}_{L1}\mathbf{W}_{L1}$ with $\mathbf{X}_{L2}$ (denoted as $D_{12}$) and (2) $\mathbf{X}_{L2}\textbf{W}_{L2}$ with $\mathbf{X}_{L1}$ (denoted as $D_{21}$). 
%We independently induce two different bilingual spaces -- $\mathbf{X'}_{L1} \cup \mathbf{X}_{L2}$ and $\mathbf{X'}_{L2} \cup \mathbf{X}_{L1}$---and detect mutual nearest neighbours (according to cosine similarity). 
We then augment the dictionary $D$ with mutual nearest neighbours we find: $D_{12} \cap D_{21}$.\footnote{In preliminary experiments, we achieve best performance using only a single bootstrapping iteration. Even when starting with $D$ of 1K entries, after the first iteration we find between 5K and 10K mutual nearest neighbours that yield a better mapping. In the second iteration already we obtain a much larger number of noisier (automatically generated) word translations, which do not lead to further performance gains.} 

\vspace{1.4mm}
\noindent \textbf{Discriminative Latent-Variable (\dlv).} \citet{ruder2018discriminative} augment the seed supervised lexicon through Expectation-Maximization in a latent-variable model. The source words $\{w^i_{L1}\}^K_{k=1}$ and target words $\{w^j_{L2}\}^K_{k=1}$ are seen as a fully connected bipartite weighted graph $G = (E, V_{L1} \cup V_{L2})$ with edges $E = V_{L1} \times V_{L2}$. By drawing embeddings from a Gaussian distribution and normalizing them, the weight of each edge $(i, j) \in E$ is shown to correspond to the cosine similarity between vectors. In the E-step, a maximal bipartite matching is found on the sparsified graph using the Jonker-Volgenant algorithm \cite{jonker1987shortest}. In the M-step, a better projection $W_{L1}$ is learned solving the Procrustes problem.

\vspace{1.4mm}
\noindent \textbf{Ranking-Based Optimization (\rcsls).} \newcite{joulin2018loss} optimize the projection matrix $\mathbf{W}_{L1}$ by maximizing the cross-domain similarity local scaling \cite[CSLS;][]{conneau2018word} score, instead of minimizing the Euclidean distances between projection of $\mathbf{X}_S$ and $\mathbf{X}_T$. %However, instead of minimizing the quadratic loss (i.e., the Euclidean distance) from Eq.~\eqref{eq:linmap_opt}, they opt for directly optimizing the cross-domain similarity local scaling \cite[CSLS;][]{conneau2018word} score, 
CSLS is a modification of cosine similarity commonly used for BLI inference. Let $r(\mathbf{x}^k_{L1}\mathbf{W}, \textbf{X}_{L2})$ be the average cosine similarity of the projected vector $\mathbf{x}^k_{L1}\mathbf{W}$ with its $N$ nearest neighbors from $\textbf{X}_{L2}$. Similarly, let $r(\mathbf{x}^k_{L2}, \textbf{X}_{L1}\mathbf{W})$ be the average cosine similarity of the target space vector $\mathbf{x}^k_{L2}$ with its $N$ nearest neighbors from the projected source space $\textbf{X}_{L1}\mathbf{W}$. By relaxing the constraint that $\mathbf{W}_{L1}$ is orthogonal,
%is relaxed and monolingual vectors from $\mathbf{X}_{L1}$ and $\mathbf{X}_{L2}$ are $\ell_2$-normalized,
 maximization of relaxed CSLS (dubbed \rcsls) becomes a convex optimization problem:

\vspace{-2mm}
{\small
\begin{align}
    \textbf{W}_{L1} &= \argmin_{\mathbf{W}} \frac{1}{K} \sum_{\substack{\mathbf{x}^k_{L1} \in X_S \\  \mathbf{x}^k_{L2} \in X_T}}{\hspace{-1em}-2\cos{\left(\mathbf{x}^k_{L1}\mathbf{W}, \mathbf{x}^k_{L2}\right)}} \notag \\ &+ r(\mathbf{x}^k_{L1}\mathbf{W}, \textbf{X}_{L2}) + r(\mathbf{x}^k_{L2}, \textbf{X}_{L1}\mathbf{W}) 
\end{align}}%
\noindent By maximizing (R)CSLS, this model is explicitly designed to induce a cross-lingual embedding space that performs well in the BLI task.

\vspace{1.4mm}
\noindent \textbf{Adversarial Alignment (\muse).}
Several CLE models initialize a seed bilingual lexicon solely from monolingal data using Generative Adversarial Networks \cite[GANs;][]{goodfellow2014generative}, with \muse \cite{conneau2018word} being the most effective example. 
%A GAN consists of a generator and a discriminator, where 
The generator component in \muse is again the linear map $\mathbf{W}_{L1}$. 
%Assuming topological similarity between two monolingual vector spaces, 
\muse aims to improve the generator mapping $\mathbf{W}_{L1}$ such that the discriminator (a binary classifier implemented as multi-layer perceptron) fails to distinguish between the vector sampled from the target language distribution 
$\mathbf{X}_{L2}$ and the projected vector $\mathbf{X}_{L1}\mathbf{W}_{L1}$. 
%The discriminator is a binary classifier implemented as a multi-layer perceptron. The model parameters are optimized using SGD by alternating between updating the discriminator's and the generator's parameters.
\muse then improves the GAN-induced mapping, through a refinement step very similar to the \procb procedure. 
%$\mathbf{W}$ is used to build a small seed dictionary of frequent words that are mutual nearest neighbours and a new projection matrix is then induced by solving the Orthogonal Procrustes problem from Eq.~\eqref{eq:proc} on the full vocabulary. This step can be applied iteratively. CSLS is used instead of the cosine similarity. 
\muse strongly assumes approximate isomorphism of monolingual spaces \cite{sogaard2018on}, often leading to poor GAN-based initialization, especially for pairs of distant languages.

\vspace{1.4mm}
\noindent \textbf{Heuristic Alignment (\vecmap).}
%To remedy for the instability issue of \muse, 
\newcite{artetxe2018robust} induce the initial seed lexicon $D$ using a heuristic that  assumes that for a pair of translations, their monolingual similarity vectors (i.e, vectors of, e.g., cosine or CSLS, similarities with all words in L1/L2) are (approximately) equal. Seed translations are generated by finding nearest neighbors based on the similarity between monolingual similarity distributions of words. Next, they employ a self-learning bootstrapping procedure similar to \muse.
%, based on solving the Procrustes problem from Eq.~\eqref{eq:proc} in each step. 
%The initial dictionary $D$ is noisy, but has been empirically shown (on BLI only!) to result in a more robust self-learning method than the GAN-initialized \muse.
In addition to the different initialization, \vecmap critically relies on a number of empirically motivated enhancements that ensure its robustness. It adopts both multi-step pre-processing consisting of unit length normalization, mean centering, and ZCA whitening \cite{Bell1997} as well several post-processing steps: cross-correlational re-weighting, de-whitening, and dimensionality reduction \cite{artetxe2018generalizing}. 
%For more details on these steps, we refer the reader to the original paper \cite{artetxe2018generalizing}. 
%Moreover, the self-learning method relies on frequency-based vocabulary cutoff (i.e., only the 20,000 most frequent words are considered in both languages), uses CSLS instead of cosine, and retains only mutual nearest neighbours. 
What is more, \vecmap critically relies on \textit{stochastic} dictionary induction: at each iteration, some elements in the similarity matrix are set to $0$ based on the probability that varies across iterations. This enables the model to escape poor local optima.

\vspace{1.4mm}
\noindent \textbf{Iterative Closest Point Model (\icp).} The unsupervised model of \newcite{hoshen2018nonadversarial} induces initial seed dictionary $D$ by projecting vectors of the $N$ most frequent words from both monolingual spaces to a lower-dimensional space with PCA. They then search for an optimal alignment between L1 words (vectors $\mathbf{x}^i_{1}$) and L2 words (vectors $\mathbf{x}^j_{2}$), assuming linear projections $\mathbf{W}_{1}$ and $\mathbf{W}_{2}$. Let $f_{1}(i)$ (vice versa $f_{2}(j)$) denote the index of the L2 word (vice versa L1 word) to which $\mathbf{x}^i_{1}$ (vice versa $\mathbf{x}^j_{2}$) is aligned. The optimization task is to find such translation matrices $\mathbf{W}_{1}$ and $\mathbf{W}_{2}$ that minimize the sum of Euclidean distances between optimally aligned vectors. They use Iterative Closest Point, a two-step iterative optimization algorithm that first fixes translation matrices $\mathbf{W}_{1}$ and $\mathbf{W}_{2}$ to find the optimal alignments and then use those alignments to update the translation matrices by minimizing: %the following objective:  

\vspace{-1.5mm}
{\footnotesize \begin{align}
\sum_i \| \mathbf{x}^{i}_{1}\mathbf{W}_{1} - \mathbf{x}^{f_{1}(i)}_{2} \|&+ \sum_j \| \mathbf{x}^{j}_{2}\mathbf{W}_{2} - \mathbf{x}^{f_{2}(j)}_{1} \| +  \notag \\ 
\lambda\sum_i \| \mathbf{x}^{i}_{1} - \mathbf{x}^{i}_{1}\mathbf{W}_{1}\mathbf{W}_{2} \| &+ \lambda\sum_j \| \mathbf{x}^{j}_{2} - \mathbf{x}^{j}_{2}\mathbf{W}_{2}\mathbf{W}_{1} \|.
\end{align}}%
\noindent The second-line terms are cyclical consistency constraints forcing vectors round-projected to the other language space and back to remain unchanged.
%, with $\lambda$ regulating their contribution.
%Next, they induece new projection matrices using final alignments between the $N$ most frequent words from both vocabularie. 
Next, they employ a dictionary bootstraping procedure based on mutual nearest neighbours (similar to \procb and \muse) and produce the final projections by solving the Procrustes problem. 
%Third, the translation matrices are further trained using only mutual nearest neighbours between the first 7,500 words in both vocabularies. Finally, the projections are fine-tuned by again solving the Procrustes problem on the full vocabulary.
%using mutual nearest neighbours obtained from the previous step for supervision.

\vspace{1.4mm}
\noindent \textbf{Gromov-Wasserstein Alignment Model (GWA).} 
Observing that word embedding models employ metric recovery algorithms, \citet{alvarez2018gromov} cast CLE induction as an optimal transport problem based on the Gromov-Wasserstein distance (with cosine distance as a cost measure). They first compute intra-language costs $\mathbf{C}_{L1} = \text{cos}(\mathbf{X}_{L1}, \mathbf{X}_{L1})$ and $\mathbf{C}_{L2} = \text{cos}(\mathbf{X}_{L2}, \mathbf{X}_{L2})$ as well as inter-language similarities $\mathbf{C}_{12} = \mathbf{C}^2_{L1} \mathbf{p} \mathbb{1}^\top_m + \mathbb{1}_n \mathbf{q} (\mathbf{C}^2_{L2})^\top$, with $\mathbf{p}$ and $\mathbf{q}$ as uniform probability distributions over $V_{L1}$ and $V_{L2}$, respectively. They then induce the initial projections by solving the Gromov-Wasserstein optimal transport problem with a fast iterative algorithm \cite{peyre2016gromov}, 
%that iteratively (1) computes a pseudo-cost matrix $\hat{\mathbf{C}}_\Gamma = \mathbf{C}_{12} - 2 \mathbf{C}_{L1} \Gamma \mathbf{C}_{L2}^\top$ where $\Gamma$ is a row- and column-stochastic alignment matrix, containing probabilities for source and target words being translation pairs and (2) solves the traditional optimal transport problem with the Sinkhorn-Knopp algorithm \cite{cuturi2013sinkhorn}, which $ that 
iteratively updating parameter vectors $\mathbf{a}$ and $\mathbf{b}$:

{\footnotesize
\begin{equation*}
\mathbf{a} = \mathbf{p} \oslash \mathbf{K} \mathbf{b}, \: \: \: \mathbf{b} = \mathbf{q} \oslash \mathbf{K}^\top \mathbf{a},
\end{equation*}}
\noindent where $\oslash$ is element-wise division and $\mathbf{K} = \exp(-\hat{\mathbf{C}}_\Gamma / \lambda )$ (with $\hat{\mathbf{C}}_\Gamma = \mathbf{C}_{12} - 2 \mathbf{C}_{L1} \Gamma \mathbf{C}_{L2}^\top$).
%, with $\lambda$ as the entropy regularization factor. 
The alignment matrix $\Gamma = \text{diag}(\mathbf{a})\, \mathbf{K}\, \text{diag}(\mathbf{b})$ is recomputed in each iteration. The final projection $\mathbf{W}_{L1}$ is (again!) obtained by solving Procrustes using the final alignments in $\Gamma$ as supervision. 

In sum, our brief overview points to the main high-level (dis)similarities of all projection-based CLE models: while they differ in the way the initial seed lexicon is extracted, most models (with \cca, \proc, and \rcsls as exceptions) are based on self-learning procedures that repeatedly solve the Procrustes problem from Eq.~\eqref{eq:proc}, typically on the trimmed vocabulary. In the final step, the fine-tuned linear map is applied on the full vocabulary.
%A similar approach has been proposed by \newcite{Grave:2018arxiv}.
\section{Bilingual Lexicon Induction}

Bilingual lexicon induction has become the \textit{de facto} standard evaluation task for projection-based CLE models. Given a shared CLE space, the task is to retrieve target language translations for a (test) set of source language words. A typical BLI evaluation in the recent literature reports comparisons with the well-known \muse model on a few language pairs, all of which involve English as one of the languages. A comprehensive comparative BLI evaluation of a wider set of models conducted on a larger set of language pairs is still missing. In our evaluation, we include a much larger set of language pairs, including pairs in which neither of the languages is English.\footnote{To the best of our knowledge, all language pairs in existing BLI evaluations involved English as one of the languages (either source or target language)---with the exception of Estonian--Finnish used by \newcite{sogaard2018on}.} Furthermore, to allow both for a fair comparison between supervised models and for performance comparisons between different language pairs, we create training and evaluation dictionaries that are \textit{fully aligned} across all evaluated language pairs. Finally, we also discuss other choices in existing BLI evaluations which are currently taken for granted: e.g., (in)appropriate evaluation metrics and lack of significance testing.

\vspace{1.4mm}
\noindent \textbf{Language Pairs.} Our evaluation comprises eight languages: Croatian (\hr), English (\en), Finnish (\fin), French (\fr), German (\de), Italian (\ita), Russian (\ru), and Turkish (\tr). For diversity, we selected two languages from three different Indo-European branches---Germanic (\en, \de), Romance (\fr, \ita), and Slavic (\hr, \ru)---as well as two non-Indo-European languages (\fin~from the Uralic family, and \tr~from the Turkic family). Creating all possible pairs between these eight languages results in a total of 28 language pairs under evaluation. 

\vspace{1.4mm}
\noindent \textbf{Monolingual Embeddings.} Following prior work, we use 300-dimensional fastText embeddings \cite{Bojanowski:2017tacl},\footnote{\url{https://fasttext.cc/docs/en/crawl-vectors.html}} pretrained on full Wikipedias of each language. We trim all vocabularies to the 200K most frequent words.

\vspace{1.4mm}
\noindent \textbf{Translation Dictionaries.} We automatically created translation dictionaries using Google Translate similar to prior work \cite{conneau2018word}. We selected the 20K most frequent English words and automatically translated them to the other seven languages. We retained only tuples for which all translations were unigrams found in vocabularies of their respective monolingual embedding spaces, leaving us with $\approx$7K tuples. We reserved 5K tuples created from the more frequent English words for training, and the remaining 2K tuples as test dictionaries. We also created two smaller training dictionaries, by selecting tuples corresponding to the 1K and 3K most frequent English words.

\vspace{1.4mm}
\noindent \textbf{Evaluation Measures and Significance.} BLI is cast as a ranking (i.e., retrieval) task; existing BLI evaluations employ $P@k$ (precision at rank $k$, commonly with $k \in \{1, 5, 10\}$) for evaluation. We advocate the use of mean average precision (MAP) instead.\footnote{In this setup with only one correct translation for each query, MAP is equivalent to mean reciprocal rank (MRR).} While correlated with $P@k$, MAP is more informative: unlike MAP, $P@k$ treats all models that rank the correct translation below $k$ equally.\footnote{For instance, when relying on $P@5$, a model that ranks the correct translation at rank 6 is equally penalized as the model that ranks it at rank 200K.} 

%% This is a minor detail
%% For simplicity, we rely on the cosine similarity, but the same relative trends are observed with the more sophisticated CSLS similarity metric.

%% (IV, Removed this, sounds trite and non-substantial)
%% With BLI test sets commonly containing only a few thousand instances, $P@k$ may lead to erroneous conclusions about relative BLI performance between models. 

The limited size of BLI test sets warrants statistical significance testing. Yet, a common case in BLI evaluation is to declare performance gains based purely on (rather limited) numeric performance increases (e.g., below 1-point $P@1$), without any significance analysis. We test the significance of BLI results by applying the two-tailed t-test with the Bonferroni correction \cite{Dror:2018acl}. %to lists of integer rankings produced by two models on the test dictionaries.\footnote{We apply the Bonferroni correction to account for testing multiple hypotheses.}        

\subsection{Results and Discussion}

%% (IV, MOVED TO TABLE CAPTION)
%%: the first column (All LPs) displays averages over all 28 language pairs; the second column (Filt.~LPs) shows averages only over language pairs for which \textit{all} models yielded at least one successful run\footnote{All unsupervised CLE models except \vecmap displayed a high level of instability being unable to produce a successful run for many language pairs. We treat a run as successful if it yields a MAP performance $\geq$ 0.05.}; and the third column (Succ.~LPs) shows the number of language pairs for which we obtained a successful run.   

\setlength{\tabcolsep}{7pt}
\begin{table}[t]
\centering
{\footnotesize
\begin{tabularx}{\linewidth}{l l c c c}
\toprule 
Model & Dict & All LPs & Filt.~LPs & Succ.~LPs \\ \midrule
\multicolumn{4}{l}{\textit{Supervised}} \\ \midrule
\cca & 1K & .289 & .404 & 28/28\\
\cca & 3K & .378 & .482 & 28/28 \\
\cca & 5K & .400 & .498 & 28/28 \\ \hdashline
\proc & 1K & .299 & .411 & 28/28 \\
\proc & 3K & .384 & .487 & 28/28 \\ 
\proc & 5K & .405 & .503 & 28/28 \\ \hdashline
\procb & 1K & .379 & .485 & 28/28 \\
\procb & 3K & .398 & .497 & 28/28 \\ \hdashline
\dlv & 1K & .289 & .400 & 28/28 \\
\dlv & 3K & .381 & .484 & 28/28 \\
\dlv & 5K & .403 & .501 & 28/28 \\ \hdashline
\rcsls & 1K & .331 & .441 & 28/28 \\
\rcsls & 3K & .415 & .511 & 28/28 \\
\rcsls & 5K & .437 & .527 & 28/28 \\ \midrule
\multicolumn{4}{l}{\textit{Unupervised}} \\ \midrule
\vecmap & -- & .375 & .471 & 28/28 \\ 
\muse & -- & .183 & .458 & {\color{red}{13}}/28 \\ 
\icp & -- & .253 & .424 & {\color{red}{22}}/28 \\ 
\gwa & -- & .137 & .345 & {\color{red}{15}}/28 \\ 
\bottomrule
\end{tabularx}
}
\vspace{-2mm}
\caption{Summary of BLI performance (MAP). \textbf{All LPs}: average scores over all 28 language pairs; \textbf{Filt.~LP}: average scores only over language pairs for which \textit{all} models in evaluation yield at least one successful run; \textbf{Succ.~LPs}: the number of language pairs for which we obtained at least one successful run. A run is considered \textit{successful} if MAP $\geq$ 0.05.}
\label{tbl:overall}
\vspace{-3mm}
\end{table}

Table \ref{tbl:overall} summarizes BLI performance over multiple language pairs. \rcsls~\cite{joulin2018loss} displays the strongest BLI performance. This is not a surprising finding given that its learning objective is tailored for BLI in particular. \rcsls outperforms other supervised models (\cca, \proc, and \dlv) trained with exactly the same dictionaries. \newcite{smith2017offline} already suggested that \cca and \proc display similar performance. We confirm that their performance, as well as the performance of \dlv, are statistically indistinguishable, even at $\alpha=0.1$. Our bootstrapping \procb~approach, significantly boosts the performance of \proc~when using a small translation dictionary of 1K pairs ($p<0.01$). For the same 1K training dictionary, \procb~significantly outperforms \rcsls~as well. Interestingly, for each supervised model, training on 5K pairs does not significantly outscore the variant trained on 3K pairs, whereas training on 3K or 5K pairs does significantly outperform the 1K variants. This confirms a finding from prior work \cite{vulic2016on} suggesting that no significant improvements are to be expected from training linear maps on dictionaries larger than 5K entries.   

The results highlight \vecmap~\cite{artetxe2018robust} as the most robust choice among all unsupervised models: besides being the only model to produce successful runs consistently for all language pairs, it also substantially outperforms all other unsupervised models. It is also the most effective unsupervised model on the subset of language pairs for which the other models produce successful runs. However, \vecmap~is still significantly outperformed ($p\leq0.0002$) by \procb trained on 1K translation pairs and by all supervised models trained on 3K and 5K word pairs. These findings challenge unintuitive claims from recent work \cite{artetxe2018robust,hoshen2018nonadversarial,alvarez2018gromov} that the unsupervised CLE models are competitive to or even surpass supervised CLE models in the BLI task. 

\setlength{\tabcolsep}{6.5pt}
\begin{table*}[t]
\centering
\def\arraystretch{0.93}
{\footnotesize
\begin{tabularx}{\linewidth}{l l cc cc cc cc cc}
\toprule 
Model & Dict & \en--\de & \ita--\fr & \hr--\ru & \en--\hr & \de--\fin & \tr--\fr & \ru--\ita & \fin--\hr & \tr--\hr & \tr--\ru \\ \midrule
\proc & 1K & 0.458 & 0.615 & 0.269 & 0.225 & 0.264 & 0.215 & 0.360 & 0.187 & 0.148 & 0.168 \\
\proc & 5K & 0.544 & 0.669 & 0.372 & 0.336 & 0.359 & 0.338 & 0.474 & 0.294 & 0.259 & 0.290 \\
\procb & 1K & 0.521 & 0.665 & 0.348 & 0.296 & 0.354 & 0.305 & 0.466 & 0.263 & 0.210 & 0.230 \\
\rcsls & 1K & 0.501 & 0.637 & 0.291 & 0.267 & 0.288 & 0.247 & 0.383 & 0.214 & 0.170 & 0.191 \\
\rcsls & 5K & 0.580 & 0.682 & 0.404 & 0.375 & 0.395 & 0.375 & 0.491 & 0.321 & 0.285 & 0.324 \\ \midrule
\vecmap & -- & 0.521 & 0.667 & 0.376 & 0.268 & 0.302 & 0.341 & 0.463 & 0.280 & 0.223 & 0.200 \\ \midrule
\textbf{Average} & -- & 0.520 & 0.656 & 0.343 & 0.294 & 0.327 & 0.304 & 0.440 & 0.260 & 0.216 & 0.234 \\
\bottomrule
\end{tabularx}
}
\vspace{-2mm}
\caption{BLI performance (MAP) with a selection of models on a subset of evaluated language pairs.}
\vspace{-2.5mm}
\label{tbl:bli_lps}
\end{table*}

Table~\ref{tbl:bli_lps} shows the scores for a subset of 10 language pairs using a subset of models from Table~\ref{tbl:overall}.\footnote{We provide full BLI results for all 28 language pairs and all models in the supplemental material (see appendix).} As expected, all models work reasonably well for major languages---this is most likely due to a higher quality of the respective monolingual embeddings, which are pre-trained on much larger corpora.\footnote{For instance, \en~Wikipedia is approximately 3 times larger than \de~and \ru~Wikipedias, 19 times larger than \fin~Wikipedia and 46 times larger than \hr~Wikipedia.} Language proximity also plays a critical role: on average, models achieve better BLI performance for languages from the same family (e.g., compare the results of \hr--\ru~vs. \hr--\en). 

The gap between the best-performing supervised model (\rcsls) and the best-performing unsupervised model (\vecmap) is more pronounced for cross-family language pairs, especially those consisting of one Germanic and one Slavic or non-Indo-European language (e.g., 19 points MAP difference for \en--\ru, 14 for \de--\ru~and \en--\fin, 10 points for \en--\tr~and \en--\hr, 9 points for \de--\fin~and \en--\hr). We suspect that this is due to heuristics based on intra-language similarity distributions, employed by \newcite{artetxe2018robust} to induce an initial translation dictionary, being less effective the more distant the languages in the pair are.  

\section{Downstream Evaluation}
We now test projection-based CLE models in extrinsic cross-lingual applications, moving beyond the limiting BLI evaluation. We select three diverse cross-lingual downstream tasks: \textbf{1)} cross-lingual transfer for natural language inference (XNLI), a sentence-level language understanding task; \textbf{2)} cross-lingual document classification (CLDC); which requires only shallow (i.e., lexical-level or topical) meaning modeling, and \textbf{3)} cross-lingual information retrieval (CLIR), an unsupervised ranking task that relies on detecting coarser semantic relatedness.

\subsection{Natural Language Inference}

Large training corpora for NLI exist only in English \cite{bowman2015large,williams2018broad}. Recently, \newcite{conneau2018xnli} released a multilingual XNLI corpus created by translating dev and test sets of the MultiNLI corpus \cite{williams2018broad} to 15 languages, hoping to foster research on cross-lingual sentence understanding. 

\vspace{1.4mm}
\noindent \textbf{Evaluation Setup.} 
The multilingual XNLI corpus covers 5 out of our 8 languages used previously for BLI: \en, \de, \fr, \ru, and \tr. Our XNLI evaluation setup is straightforward: we train a well-known and robust neural NLI model, Enhanced Sequential Inference Model \cite[ESIM;][]{chen2017enhanced} as implemented by \citet{williams2018broad},\footnote{Since our aim is to compare different bilingual spaces---input vectors for ESIM, kept fixed during training---we simply use the default ESIM hyper-parameter configuration.} on the large English MultiNLI corpus, using \en word embeddings from a shared \en--L2 (L2 $\in$ \{\de, \fr, \ru, \tr\}) embedding space. We then evaluate the model on the L2 XNLI test set by feeding L2 embeddings from the shared space.\footnote{Note that one is able to use exactly the same ESIM model for all direct-projection CLE models (i.e., those not changing L2 vectors, $\mathbf{W}_{L2} = I$) by using \en~as L2, since different CLE models only produce different L1 vectors after projection. Exceptions are \vecmap~and \dlv~models which transform the embeddings in L2 as well: for these models we had to train a different ESIM model for each language pair.}\textsuperscript{,}\footnote{The goal of this evaluation is not to compete with current state-of-the-art systems for (X)NLI \cite{artetxe2018massively,lample2019cross}, but rather to provide means to analyze properties and relative performance of diverse CLE architectures in a downstream understanding task.}

\vspace{1.4mm}
\noindent \textbf{Results and Discussion.} 
XNLI accuracy scores are summarized in Table~\ref{tbl:nli}. 
\setlength{\tabcolsep}{2.7pt}
\begin{table}[t]
\centering
{\footnotesize
\begin{tabularx}{\linewidth}{l l ccccc}
\toprule 
Model & Dict & \en--\de & \en--\fr & \en--\tr & \en--\ru & Avg \\ \midrule
\multicolumn{4}{l}{\textit{Supervised}} \\ \midrule
\proc & 1K & 0.561 & 0.504 & 0.534 & 0.544 & 0.536 \\
\proc & 5K & 0.607 & 0.534 & 0.568 & 0.585 & \textbf{0.574} \\ 
\procb & 1K & 0.613 & 0.543 & 0.568 & 0.593 & \textbf{0.579} \\ 
\procb & 3K & 0.615 & 0.532 & 0.573 & 0.599 & \textbf{0.580} \\
\dlv & 5K & 0.614 & 0.556 & 0.536 & 0.579 & \textbf{0.571} \\ 
\rcsls & 1K & 0.376 & 0.357 &  0.387 & 0.378 & 0.374 \\
\rcsls & 5K & 0.390 & 0.363 & 0.387 & 0.399 & 0.385 \\ \midrule
\multicolumn{4}{l}{\textit{Unupervised}} \\ \midrule
\vecmap & -- & 0.604 & 0.613 & 0.534 & 0.574 & \textbf{0.581} \\ 
\muse & -- & 0.611 & 0.536 & 0.359* & 0.363* & 0.467 \\ 
\icp & -- & 0.580 & 0.510 & 0.400* & 0.572 & 0.516 \\
\gwa & -- & 0.427* & 0.383* & 0.359* & 0.376* & 0.386 \\
\bottomrule
\end{tabularx}
}
\vspace{-1.5mm}
\caption{XNLI performance (test set accuracy). Bold: highest scores, with mutually insignificant differences according to the non-parametric shuffling test \cite{yeh2000more}. Asterisks denote language pairs for which CLE models could not yield successful runs in the BLI task.}
\label{tbl:nli}
\vspace{-2mm}
\end{table}
%
% horrible correlation between NLI and BLI -- measure on all language pairs and all models 
%
The mismatch between BLI and XNLI performance is most obvious for \rcsls. While \rcsls~is the best-performing model in the BLI task, it shows suboptimal performance on XNLI across the board. This indicates that specializing/overfitting CLE spaces to word translation may seriously hurt cross-lingual transfer for language understanding tasks. As the second indication of the mismatch, the unsupervised \vecmap~model, outperformed by supervised models on BLI, performs on par with \proc and \procb on XNLI. Finally, there are significant differences between BLI and XNLI performance across language pairs for most models---while we observe drastically better BLI performance for \en--\de~and \en--\fr~compared to \en--\ru~and especially \en--\tr, XNLI performance of most models for \en--\ru~and \en--\tr~is better than for \en--\fr~and close to that for \en--\de. While this can also be an artefact of the XNLI dataset creation, we support these individual observations by measuring an overall correlation (Spearman's $\rho$) of only 0.13 between corresponding BLI and XNLI scores over individual language-pair scores (for all models).

The \proc model performs significantly better on XNLI when trained on 5K pairs than with 1K pairs, and this is consistent with BLI results. However, we show that we can reach the same performance level using 1K pairs and our proposed \procb bootstrapping scheme. \vecmap is again the most robust and most effective unsupervised model, but it is outperformed by the \procb model on more distant language pairs: \en--\tr~and \en--\ru. %(i.e., by training the \procb model on 1K pairs).

% RCSLS sucks (and it yields very successful runs for BLI), Artetxe is good
% EN-FR which is better than EN-RU and EN-TR in BLI is worse in NLI
% Proc-B trained on 1K performs on par in CLNLI with Proc 5K (and significantly better than Proc 1K) 

\subsection{Document Classification}

%% (IV, This was said before)
%Unlike NLI, document classification typically does not require deep language understanding and shallow semantic representations often suffice for solid performance.

\noindent \textbf{Evaluation Setup.} 
We next evaluate CLEs on the cross-lingual document classification (CLDC) task. We use the TED CLDC corpus compiled by \citet{hermann2014multilingual}. It includes 15 different topics and 12 language pairs (with \en as one of the languages in all pairs). A binary classifier is trained and evaluated for each topic and each language pair, using predetermined train and test splits. The intersection between TED languages and our BLI languages results in five CLDC evaluation pairs: \en--\de, \en--\fr, \en--\ita, \en--\ru, and \en--\tr. Since our goal is to compare different CLEs and isolate their contribution (i.e., we do not aim to match state-of-the-art on TED CLDC), for the sake of simplicity we rely on a simple light-weight CNN-based classifier in all CLDC experimental runs.\footnote{We implemented a convolutional network with a single 1-D convolutional layer (with number of parameters fixed to 8 filters for each of the sizes \{2, 3, 4, 5\}) and a max pooling layer, coupled with a softmax classifier. We minimized the standard negative log-likelihood loss using the Adam algorithm \cite{kingma2014adam}.} 

\vspace{1.4mm}
\noindent \textbf{Results and Discussion.} The results in terms of $F_1$ scores micro-averaged over 12 classes are shown in Table~\ref{tbl:cldc}.  
\setlength{\tabcolsep}{3.2pt}
\begin{table}[t]
\centering
{\footnotesize
\begin{tabularx}{\linewidth}{l l cccccc}
\toprule 
Model & Dict & \de & \fr & \ita & \ru & \tr & Avg \\ \midrule
\multicolumn{8}{l}{\textit{Supervised}} \\ \midrule
\proc & 1K & .250 & .107 & .158 & .127 & .309 & .190 \\
\proc & 5K & .345 & .239 & .310 & .251 & .190 & .267 \\
\procb & 1K & .374 & .182 & .205 & .243 & .254 & .251 \\
\procb & 3K & .352 & .210 & .218 & .186 & .310 & .255 \\
\dlv & 5K & .299 & .175 & .234 & .375 & .208 & .258 \\ 
\rcsls & 1K & .557 & .550 & .516 & .466 & .419 & \textbf{.501} \\
\rcsls & 5K & .588 & .540 & .451 & .527 & .447 & \textbf{.510} \\ \midrule
\multicolumn{8}{l}{\textit{Unupervised}} \\ \midrule
\vecmap & -- & .433 & .316 & .333 & .504 & .439 & \textbf{.405} \\ 
\muse & -- & .288 & .223 & .198 & .226* & .264* & .240 \\ 
\icp & -- & .492 & .254 & .457 & .362 & .175* & .348 \\
\gwa & -- & .180* & .209* & .206* & .151* & .173* & .184 \\
\bottomrule
\end{tabularx}
}
\vspace{-1.5mm}
\caption{CLDC performance (micro-averaged $F_1$ scores); cross-lingual transfer \en--X. Numbers in bold denote the best scores in the model group. Asterisks denote language pairs for which CLE models did not yield successful runs in the BLI task.}
\label{tbl:cldc}
\vspace{-1.5mm}
\end{table}
%
%Overall, CLDC scores are again poorly correlated with BLI results: Spearman's $\rho$ correlation is only 0.17 measured across all CLE models and all five language pairs. However, 
In contrast to XNLI, \rcsls, the best-performing BLI model, also obtains peak scores on CLDC, with a wide margin to all other models. It significantly outperforms the unsupervised \vecmap~model, which in turn significantly outperforms all other supervised models. Somewhat surprisingly, supervised Procrustes-based models (\proc, \procb, and \dlv) that performed strongly on both BLI and XNLI display very weak performance on CLDC: this calls for further analyses in future work.

\setlength{\tabcolsep}{7pt}
\begin{table*}[!t]
\centering
{\footnotesize
\begin{tabularx}{\linewidth}{l l cccccccccc}
\toprule 
Model & Dict & \de-\fin & \de-\ita & \de-\ru & \en-\de & \en-\fin & \en-\ita & \en-\ru & \fin-\ita & \fin-\ru & Avg \\ \midrule
\multicolumn{8}{l}{\textit{Supervised}} \\ \midrule
\proc & 1K & 0.147 & 0.155 & 0.098 & 0.175 & 0.101 & 0.210 & 0.104 & 0.113 & 0.096 & 0.133 \\
\proc & 5K & 0.255 & 0.212 & 0.152 & 0.261 & 0.200 & 0.240 & 0.152 & 0.149 & 0.146 & 0.196 \\
\procb & 1K & 0.294 & 0.230 & 0.155 & 0.288 & 0.258 & 0.265 & 0.166 & 0.151 & 0.136 & \textbf{0.216} \\
\procb & 3K & 0.305 & 0.232 & 0.143 & 0.238 & 0.267 & 0.269 & 0.150 & 0.163 & 0.170 & \textbf{0.215} \\
\dlv & 5K & 0.255 & 0.210 & 0.155 & 0.260 & 0.206 & 0.240 & 0.151 & 0.147 & 0.147 & 0.197 \\
\rcsls & 1K & 0.114 & 0.133 & 0.077 & 0.163 & 0.063 & 0.163 & 0.106 & 0.074 & 0.069 & 0.107 \\
\rcsls & 5K & 0.196 & 0.189 & 0.122 & 0.237 & 0.127 & 0.210 & 0.133 & 0.130 & 0.113 & 0.162 \\ \midrule
\multicolumn{8}{l}{\textit{Unupervised}} \\ \midrule
\vecmap & -- & 0.240 & 0.129 & 0.162 & 0.200 & 0.150 & 0.201 & 0.104 & 0.096 & 0.109 & 0.155 \\ 
\muse & -- & 0.001* & 0.210 & 0.195 & 0.280 & 0.000* & 0.272 & 0.002* & 0.002* & 0.001* & 0.107 \\ 
\icp & -- & 0.252 & 0.170 & 0.167 & 0.230 & 0.230 & 0.231 & 0.119 & 0.117 & 0.124 & \textbf{0.182} \\
\gwa & -- & 0.218 & 0.139 & 0.149 & 0.013 & 0.005* & 0.007* & 0.005* & 0.058 & 0.052 & 0.072 \\
\bottomrule
\end{tabularx}
}
\vspace{-2mm}
\caption{CLIR performance (MAP) of CLE models (the first language in each column is the query language, the second is the language of the document collection). Numbers in bold denote the best scores in the model group. Asterisks denote language pairs for which CLE models did not yield successful runs in BLI evaluation.}
\label{tbl:clir}
\vspace{-2mm}
\end{table*}

\subsection{Information Retrieval}

Finally, we investigate the usefulness of CLE models in cross-lingual information retrieval. Unlike XNLI and CLDC, we perform CLIR in an unsupervised fashion by comparing aggregate semantic representations of queries in one language with aggregate semantic representations of documents in another language. We perceive document retrieval to require more language understanding than CLDC (where we capture relevant n-grams) and less language understanding than XNLI (which requires modeling of subtle nuances in sentence meaning).   

\vspace{1.4mm}
\noindent \textbf{Evaluation Setup.}
We employ a simple yet effective unsupervised CLIR model from \newcite{litschko2018unsupervised}\footnote{The model is dubbed \textsc{BWE-Agg} in the original paper.} that (1) builds query and document representations as weighted averages of word embeddings from a shared cross-lingual space and (2) computes the relevance score simply as the cosine similarity between aggregate query and document vectors. We evaluate the models on the standard test collections from the CLEF 2000-2003 ad-hoc retrieval Test Suite,\footnote{\url{http://catalog.elra.info/en-us/repository/browse/ELRA-E0008/}} again using the intersection with BLI languages. We preprocessed the test collections by lowercasing queries and documents and removing punctuation and single-character tokens.   

%% We consider all query- and document language combinations for which both languages are among the eight languages we considered in our BLI evaluation. 
%% (because the entries in corresponding fastText embedding vocabularies are lowercased) 

\paragraph{Results and Discussion.} CLIR MAP scores for nine language pairs are summarized in Table~\ref{tbl:clir}. The supervised Procrustes-based methods (\proc, \procb, and \dlv) appear to have an edge in CLIR, with our bootstrapping \procb model outperforming all other CLE methods.\footnote{Similarly to the BLI evaluation, we test the significance by applying a Student's t-test on two lists of ranks of relevant documents (concatenated across all test collections), produced by two models under comparison. Even with Bonferroni correction for multiple tests, \procb significantly outperforms all other CLE models at $\alpha = 0.05$.} Contrary to other downstream tasks, \vecmap is not the best-performing unsupervised model on CLIR -- \icp significantly outperforms \vecmap (and also \muse and \gwa). The best-performing BLI model, \rcsls, displays poor CLIR performance, significantly worse than the simple \proc model. 

%%Like XNLI performance, tailoring the objective of the CLE model to word translation seems to hurt retrieval performance as well.

\subsection{Further Discussion}

At first glance BLI performance shows only a weak and inconsistent correlation with results in downstream tasks. The behaviour of \rcsls is especially peculiar: it is the best-performing BLI model and it achieves the best results on CLDC by a wide margin, but it is not at all competitive on XNLI and falls short of other supervised models in the CLIR task. Trends in downstream results for other models (i.e., excluding \rcsls) seem to roughly correspond to trends in BLI scores. 

To further investigate this, in Table~\ref{tbl:corrs} we measure correlations (in terms of Pearson's $\rho$) between aggregate task performances on BLI and three downstream tasks by considering (1) all models and (2) all models except \rcsls.\footnote{For measuring correlation between BLI and each downstream task T, we average model's BLI performance only over the language pairs included in the task T.} 
\setlength{\tabcolsep}{12pt}
\begin{table}[t]
\centering
{\footnotesize
\begin{tabularx}{\linewidth}{l ccc}
\toprule 
Models & XNLI & CLDC & CLIR \\ \midrule
All models & 0.269 & 0.390 & 0.764 \\
All w/o \rcsls & \textbf{0.951} & 0.266 & \textbf{0.910} \\
\bottomrule
\end{tabularx}
}
\vspace{-2mm}
\caption{Correlations of model-level results between BLI and each of the three downstream tasks.}
\label{tbl:corrs}
\vspace{-2mm}
\end{table}
Without \rcsls, BLI results correlate almost perfectly with the results on XNLI and CLIR, while they correlate weakly with CLDC results. %This is completely aligned with the behaviour of the \textit{RCSLS} model which performs great on CLDC but poorly on XNLI and CLIR. But what makes \textit{RCSLS} behave so differently compared to all other models?

The question is: why does \rcsls diverge from other models? All other models except \rcsls in the final step induce an orthogonal projection matrix using external or automatically induced translation dictionaries, minimizing the Euclidean distances between aligned words. In contrast, in order to maximize the CSLS similarity between aligned words, \rcsls relaxes the orthogonality condition imposed on the projection matrix $\mathbf{W}_{L1}$. This allows for distortions of the source embedding space after projection. The exact nature of these distortions and their impact on downstream performance of \rcsls require further investigation. However, these findings indicate that downstream evaluation is even more important for CLE models that learn non-orthogonal projections. For CLE models with orthogonal projections, downstream results seems to be more in line with BLI performance. 

The brief task correlation analysis is based on coarse-grained model-level aggregation. The actual selection of the strongest baseline models requires finer-grained tuning at the level of particular language pairs and evaluation tasks. However, our experiments also detect two robust baselines that should be included as indicative reference points in future research: \procb (supervised) and \vecmap (unsupervised).

% correlations between all tasks (performance)
% BLI poor predictor of downstream performance
% grade downstream tasks by "deep language understanding" and relate it to optimizing for word translation (RCSLS). 
% Argue for Proc-B, especially in downstream and for VecMap as the best unsupervised model across the board

%\input{6-Analysis.tex}
\section{Conclusion}
The rapid progress in cross-lingual word embedding (CLE) methodology is currently not matched with the adequate progress in their fair and systematic evaluation and comparative analyses. CLE models are typically evaluated on a single task only: bilingual lexicon induction (BLI), and even the BLI task includes a wide variety of evaluation setups which are not directly comparable and thus hinder our ability to correctly interpret and generalize the key results. In this work, we have made the first step towards a comprehensive evaluation of cross-lingual word embeddings. By conducting a systematic evaluation of CLE models on a large number of language pairs in the BLI task and three downstream tasks, we have shed new light on the ability of current cutting-edge CLE models to support cross-lingual NLP. In particular, we have empirically proven that the quality of CLE models is largely task-dependent and that overfitting the models to the BLI task can result in deteriorated performance in downstream tasks. We have also highlighted the most robust supervised and unsupervised CLE models and have exposed the need for reassessing existing baselines, as well as for unified and comprehensive evaluation protocols. We hope that this study will encourage future work on CLE evaluation and analysis, and also assist in guiding the development of new CLE models.

\clearpage
\bibliography{references}

\begin{thebibliography}{61}
\expandafter\ifx\csname natexlab\endcsname\relax\def\natexlab#1{#1}\fi

\bibitem[{Alaux et~al.(2019)Alaux, Grave, Cuturi, and
  Joulin}]{alaux2019unsupervised}
Jean Alaux, Edouard Grave, Marco Cuturi, and Armand Joulin. 2019.
\newblock \href {http://arxiv.org/abs/1811.01124} {Unsupervised hyperalignment
  for multilingual word embeddings}.
\newblock In \emph{Proceedings of ICLR}.

\bibitem[{Alvarez-Melis and Jaakkola(2018)}]{alvarez2018gromov}
David Alvarez-Melis and Tommi Jaakkola. 2018.
\newblock \href {http://aclweb.org/anthology/D18-1214} {{Gromov-W}asserstein
  alignment of word embedding spaces}.
\newblock In \emph{Proceedings of EMNLP}, pages 1881--1890.

\bibitem[{Ammar et~al.(2016)Ammar, Mulcaire, Tsvetkov, Lample, Dyer, and
  Smith}]{ammar2016massively}
Waleed Ammar, George Mulcaire, Yulia Tsvetkov, Guillaume Lample, Chris Dyer,
  and Noah~A Smith. 2016.
\newblock \href {https://arxiv.org/abs/1602.01925} {Massively multilingual word
  embeddings}.
\newblock \emph{arXiv preprint arXiv:1602.01925}.

\bibitem[{Artetxe et~al.(2017)Artetxe, Labaka, and
  Agirre}]{artetxe2017learning}
Mikel Artetxe, Gorka Labaka, and Eneko Agirre. 2017.
\newblock \href {http://aclweb.org/anthology/P17-1042} {Learning bilingual word
  embeddings with (almost) no bilingual data}.
\newblock In \emph{Proceedings of ACL}, pages 451--462.

\bibitem[{Artetxe et~al.(2018{\natexlab{a}})Artetxe, Labaka, and
  Agirre}]{artetxe2018generalizing}
Mikel Artetxe, Gorka Labaka, and Eneko Agirre. 2018{\natexlab{a}}.
\newblock \href
  {https://www.aaai.org/ocs/index.php/AAAI/AAAI18/paper/view/16935}
  {Generalizing and improving bilingual word embedding mappings with a
  multi-step framework of linear transformations}.
\newblock In \emph{Proceedings of AAAI}, pages 5012--5019.

\bibitem[{Artetxe et~al.(2018{\natexlab{b}})Artetxe, Labaka, and
  Agirre}]{artetxe2018robust}
Mikel Artetxe, Gorka Labaka, and Eneko Agirre. 2018{\natexlab{b}}.
\newblock \href {http://aclweb.org/anthology/P18-1073} {A robust self-learning
  method for fully unsupervised cross-lingual mappings of word embeddings}.
\newblock In \emph{Proceedings of ACL}, pages 789--798.

\bibitem[{Artetxe et~al.(2018{\natexlab{c}})Artetxe, Labaka, Agirre, and
  Cho}]{artetxe2018unsupervised}
Mikel Artetxe, Gorka Labaka, Eneko Agirre, and Kyunghyun Cho.
  2018{\natexlab{c}}.
\newblock \href {http://arxiv.org/abs/1710.11041} {Unsupervised neural machine
  translation}.
\newblock In \emph{Proceedings of ICLR}.

\bibitem[{Artetxe and Schwenk(2018)}]{artetxe2018massively}
Mikel Artetxe and Holger Schwenk. 2018.
\newblock \href {http://arxiv.org/abs/1812.10464} {Massively multilingual
  sentence embeddings for zero-shot cross-lingual transfer and beyond}.
\newblock \emph{CoRR}, abs/1812.10464.

\bibitem[{Bell and Sejnowski(1997)}]{Bell1997}
Anthony Bell and Terrence Sejnowski. 1997.
\newblock \href {http://arxiv.org/abs/9809069v1} {{The 'Independent Components'
  of Natural Scenes are Edge Filters}}.
\newblock \emph{Vision Research}.

\bibitem[{Bojanowski et~al.(2017)Bojanowski, Grave, Joulin, and
  Mikolov}]{Bojanowski:2017tacl}
Piotr Bojanowski, Edouard Grave, Armand Joulin, and Tomas Mikolov. 2017.
\newblock \href {http://arxiv.org/abs/1607.04606} {Enriching word vectors with
  subword information}.
\newblock \emph{Transactions of the ACL}, 5:135--146.

\bibitem[{Bowman et~al.(2015)Bowman, Angeli, Potts, and
  Manning}]{bowman2015large}
Samuel~R. Bowman, Gabor Angeli, Christopher Potts, and Christopher~D Manning.
  2015.
\newblock \href {http://aclweb.org/anthology/D15-1075} {A large annotated
  corpus for learning natural language inference}.
\newblock In \emph{Proceedings of EMNLP}, pages 632--642.

\bibitem[{Chen et~al.(2017)Chen, Zhu, Ling, Wei, Jiang, and
  Inkpen}]{chen2017enhanced}
Qian Chen, Xiaodan Zhu, Zhen-Hua Ling, Si~Wei, Hui Jiang, and Diana Inkpen.
  2017.
\newblock \href {http://aclweb.org/anthology/P17-1152} {Enhanced {LSTM} for
  natural language inference}.
\newblock In \emph{Proceedings of ACL}, pages 1657--1668.

\bibitem[{Chen and Cardie(2018)}]{chen2018unsupervised}
Xilun Chen and Claire Cardie. 2018.
\newblock \href {http://aclweb.org/anthology/D18-1024} {Unsupervised
  multilingual word embeddings}.
\newblock In \emph{Proceedings of EMNLP}, pages 261--270.

\bibitem[{Conneau et~al.(2018{\natexlab{a}})Conneau, Lample, Ranzato, Denoyer,
  and J{\'e}gou}]{conneau2018word}
Alexis Conneau, Guillaume Lample, Marc'Aurelio Ranzato, Ludovic Denoyer, and
  Herv{\'e} J{\'e}gou. 2018{\natexlab{a}}.
\newblock \href {https://arxiv.org/abs/1710.04087} {Word translation without
  parallel data}.
\newblock In \emph{Proceedings of ICLR}.

\bibitem[{Conneau et~al.(2018{\natexlab{b}})Conneau, Rinott, Lample, Williams,
  Bowman, Schwenk, and Stoyanov}]{conneau2018xnli}
Alexis Conneau, Ruty Rinott, Guillaume Lample, Adina Williams, Samuel Bowman,
  Holger Schwenk, and Veselin Stoyanov. 2018{\natexlab{b}}.
\newblock \href {http://aclweb.org/anthology/D18-1269} {{XNLI: E}valuating
  cross-lingual sentence representations}.
\newblock In \emph{Proceedings of EMNLP}, pages 2475--2485.

\bibitem[{Dou et~al.(2018)Dou, Zhou, and Huang}]{dou2018unsupervised}
Zi-Yi Dou, Zhi-Hao Zhou, and Shujian Huang. 2018.
\newblock \href {http://aclweb.org/anthology/D18-1062} {Unsupervised bilingual
  lexicon induction via latent variable models}.
\newblock In \emph{Proceedings of EMNLP}, pages 621--626.

\bibitem[{Doval et~al.(2018)Doval, Camacho-Collados, Espinosa~Anke, and
  Schockaert}]{doval2018improving}
Yerai Doval, Jose Camacho-Collados, Luis Espinosa~Anke, and Steven Schockaert.
  2018.
\newblock \href {http://aclweb.org/anthology/D18-1027} {Improving cross-lingual
  word embeddings by meeting in the middle}.
\newblock In \emph{Proceedings of EMNLP}, pages 294--304.

\bibitem[{Dror et~al.(2018)Dror, Baumer, Shlomov, and Reichart}]{Dror:2018acl}
Rotem Dror, Gili Baumer, Segev Shlomov, and Roi Reichart. 2018.
\newblock \href {https://aclanthology.info/papers/P18-1128/p18-1128} {The
  hitchhiker's guide to testing statistical significance in natural language
  processing}.
\newblock In \emph{Proceedings of ACL}, pages 1383--1392.

\bibitem[{Faruqui and Dyer(2014)}]{faruqui2014improving}
Manaal Faruqui and Chris Dyer. 2014.
\newblock Improving vector space word representations using multilingual
  correlation.
\newblock In \emph{Proceedings of the 14th Conference of the European Chapter
  of the Association for Computational Linguistics}, pages 462--471.

\bibitem[{Goodfellow et~al.(2014)Goodfellow, Pouget{-}Abadie, Mirza, Xu,
  Warde{-}Farley, Ozair, Courville, and Bengio}]{goodfellow2014generative}
Ian~J. Goodfellow, Jean Pouget{-}Abadie, Mehdi Mirza, Bing Xu, David
  Warde{-}Farley, Sherjil Ozair, Aaron~C. Courville, and Yoshua Bengio. 2014.
\newblock \href
  {http://papers.nips.cc/book/advances-in-neural-information-processing-systems-27-2014}
  {Generative adversarial nets}.
\newblock In \emph{Proceedings of NIPS}, pages 2672--2680.

\bibitem[{Gouws et~al.(2015)Gouws, Bengio, and Corrado}]{gouws2015bilbowa}
Stephan Gouws, Yoshua Bengio, and Greg Corrado. 2015.
\newblock \href {http://jmlr.org/proceedings/papers/v37/gouws15.html}
  {{BilBOWA: F}ast bilingual distributed representations without word
  alignments}.
\newblock In \emph{Proceedings of ICML}, pages 748--756.

\bibitem[{Grave et~al.(2018)Grave, Joulin, and Berthet}]{Grave:2018arxiv}
Edouard Grave, Armand Joulin, and Quentin Berthet. 2018.
\newblock \href {http://arxiv.org/abs/1805.11222} {Unsupervised alignment of
  embeddings with {W}asserstein procrustes}.
\newblock \emph{CoRR}, abs/1805.11222.

\bibitem[{Guo et~al.(2015)Guo, Che, Yarowsky, Wang, and Liu}]{guo2015cross}
Jiang Guo, Wanxiang Che, David Yarowsky, Haifeng Wang, and Ting Liu. 2015.
\newblock \href {http://aclweb.org/anthology/P15-1119} {Cross-lingual
  dependency parsing based on distributed representations}.
\newblock In \emph{Proceedings of ACL}, pages 1234--1244.

\bibitem[{Hauer et~al.(2017)Hauer, Nicolai, and
  Kondrak}]{hauer2017bootstrapping}
Bradley Hauer, Garrett Nicolai, and Grzegorz Kondrak. 2017.
\newblock \href {http://aclweb.org/anthology/E17-2098} {Bootstrapping
  unsupervised bilingual lexicon induction}.
\newblock In \emph{Proceedings of EACL}, pages 619--624.

\bibitem[{Hermann and Blunsom(2014)}]{hermann2014multilingual}
Karl~Moritz Hermann and Phil Blunsom. 2014.
\newblock \href {http://aclweb.org/anthology/P14-1006} {Multilingual models for
  compositional distributed semantics}.
\newblock In \emph{Proceedings of ACL}, pages 58--68.

\bibitem[{Heyman et~al.(2017)Heyman, Vuli{\'{c}}, and
  Moens}]{heyman2017bilingual}
Geert Heyman, Ivan Vuli{\'{c}}, and Marie-Francine Moens. 2017.
\newblock \href {http://aclweb.org/anthology/E17-1102} {Bilingual lexicon
  induction by learning to combine word-level and character-level
  representations}.
\newblock In \emph{Proceedings of EACL}, pages 1085--1095.

\bibitem[{Hoshen and Wolf(2018)}]{hoshen2018nonadversarial}
Yedid Hoshen and Lior Wolf. 2018.
\newblock \href {http://aclweb.org/anthology/D18-1043} {Non-adversarial
  unsupervised word translation}.
\newblock In \emph{Proceedings of EMNLP}, pages 469--478.

\bibitem[{Huang et~al.(2015)Huang, Gardner, Papalexakis, Faloutsos,
  Sidiropoulos, Mitchell, Talukdar, and Fu}]{huang2015translation}
Kejun Huang, Matt Gardner, Evangelos Papalexakis, Christos Faloutsos, Nikos
  Sidiropoulos, Tom Mitchell, Partha~P. Talukdar, and Xiao Fu. 2015.
\newblock \href {http://aclweb.org/anthology/D15-1127} {Translation invariant
  word embeddings}.
\newblock In \emph{Proceedings of EMNLP}, pages 1084--1088.

\bibitem[{Jonker and Volgenant(1987)}]{jonker1987shortest}
Roy Jonker and Anton Volgenant. 1987.
\newblock \href {https://link.springer.com/article/10.1007/BF02278710} {A
  shortest augmenting path algorithm for dense and sparse linear assignment
  problems}.
\newblock \emph{Computing}, 38(4):325--340.

\bibitem[{Joulin et~al.(2018)Joulin, Bojanowski, Mikolov, J{\'e}gou, and
  Grave}]{joulin2018loss}
Armand Joulin, Piotr Bojanowski, Tomas Mikolov, Herv{\'e} J{\'e}gou, and
  Edouard Grave. 2018.
\newblock \href {http://aclweb.org/anthology/D18-1330} {Loss in translation:
  {L}earning bilingual word mapping with a retrieval criterion}.
\newblock In \emph{Proceedings of EMNLP}, pages 2979--2984.

\bibitem[{Kim et~al.(2018)Kim, Geng, and Ney}]{kim2018improving}
Yunsu Kim, Jiahui Geng, and Hermann Ney. 2018.
\newblock \href {http://aclweb.org/anthology/D18-1101} {Improving unsupervised
  word-by-word translation with language model and denoising autoencoder}.
\newblock In \emph{Proceedings of EMNLP}, pages 862--868.

\bibitem[{Kingma and Ba(2014)}]{kingma2014adam}
Diederik~P Kingma and Jimmy Ba. 2014.
\newblock \href {https://arxiv.org/abs/1412.6980} {Adam: A method for
  stochastic optimization}.
\newblock \emph{arXiv preprint arXiv:1412.6980}.

\bibitem[{Klementiev et~al.(2012)Klementiev, Titov, and
  Bhattarai}]{klementiev2012inducing}
Alexandre Klementiev, Ivan Titov, and Binod Bhattarai. 2012.
\newblock \href {http://aclweb.org/anthology/C12-1089} {Inducing crosslingual
  distributed representations of words}.
\newblock \emph{Proceedings of COLING}, pages 1459--1474.

\bibitem[{Lample and Conneau(2019)}]{lample2019cross}
Guillaume Lample and Alexis Conneau. 2019.
\newblock \href {https://arxiv.org/abs/1901.07291} {Cross-lingual language
  model pretraining}.
\newblock \emph{CoRR}, abs/1901.07291.

\bibitem[{Lample et~al.(2018)Lample, Ott, Conneau, Denoyer, and
  Ranzato}]{lample2018phrase}
Guillaume Lample, Myle Ott, Alexis Conneau, Ludovic Denoyer, and Marc'Aurelio
  Ranzato. 2018.
\newblock \href {http://aclweb.org/anthology/D18-1549} {Phrase-based {\&}
  neural unsupervised machine translation}.
\newblock In \emph{Proceedings of EMNLP}, pages 5039--5049.

\bibitem[{Levy et~al.(2017)Levy, S{\o}gaard, and Goldberg}]{levy2017strong}
Omer Levy, Anders S{\o}gaard, and Yoav Goldberg. 2017.
\newblock \href {http://aclweb.org/anthology/E17-1072} {A strong baseline for
  learning cross-lingual word embeddings from sentence alignments}.
\newblock In \emph{Proceedings of EACL}, pages 765--774.

\bibitem[{Litschko et~al.(2018)Litschko, Glava\v{s}, Ponzetto, and
  Vuli\'{c}}]{litschko2018unsupervised}
Robert Litschko, Goran Glava\v{s}, Simone~Paolo Ponzetto, and Ivan Vuli\'{c}.
  2018.
\newblock \href {https://arxiv.org/abs/1805.00879} {Unsupervised cross-lingual
  information retrieval using monolingual data only}.
\newblock In \emph{Proceedings of SIGIR}, pages 1253--1256.

\bibitem[{Luong et~al.(2015)Luong, Pham, and Manning}]{luong2015bilingual}
Thang Luong, Hieu Pham, and Christopher~D. Manning. 2015.
\newblock \href {http://aclweb.org/anthology/W15-1521} {Bilingual word
  representations with monolingual quality in mind}.
\newblock In \emph{Proceedings of the 1st Workshop on Vector Space Modeling for
  Natural Language Processing}, pages 151--159.

\bibitem[{Mayhew et~al.(2017)Mayhew, Tsai, and Roth}]{mayhew2017cheap}
Stephen Mayhew, Chen-Tse Tsai, and Dan Roth. 2017.
\newblock \href {http://aclweb.org/anthology/D17-1269} {Cheap translation for
  cross-lingual named entity recognition}.
\newblock In \emph{Proceedings of EMNLP}, pages 2536--2545.

\bibitem[{Mikolov et~al.(2013)Mikolov, Le, and
  Sutskever}]{mikolov2013exploiting}
Tomas Mikolov, Quoc~V Le, and Ilya Sutskever. 2013.
\newblock \href {https://arxiv.org/abs/1309.4168} {Exploiting similarities
  among languages for machine translation}.
\newblock \emph{CoRR, abs/1309.4168}.

\bibitem[{Mukherjee et~al.(2018)Mukherjee, Yamada, and
  Hospedales}]{mukherjee2018learning}
Tanmoy Mukherjee, Makoto Yamada, and Timothy Hospedales. 2018.
\newblock \href {http://aclweb.org/anthology/D18-1063} {Learning unsupervised
  word translations without adversaries}.
\newblock In \emph{Proceedings of EMNLP}, pages 627--632.

\bibitem[{Nakashole(2018)}]{nakashole2018norma}
Ndapa Nakashole. 2018.
\newblock \href {http://aclweb.org/anthology/D18-1047} {{NORMA: N}eighborhood
  sensitive maps for multilingual word embeddings}.
\newblock In \emph{Proceedings of EMNLP}, pages 512--522.

\bibitem[{Peyr{\'e} et~al.(2016)Peyr{\'e}, Cuturi, and
  Solomon}]{peyre2016gromov}
Gabriel Peyr{\'e}, Marco Cuturi, and Justin Solomon. 2016.
\newblock \href {http://proceedings.mlr.press/v48/peyre16.pdf}
  {{Gromov-Wasserstein} averaging of kernel and distance matrices}.
\newblock In \emph{Proceedings of ICML}, pages 2664--2672.

\bibitem[{Ruder et~al.(2018{\natexlab{a}})Ruder, Cotterell, Kementchedjhieva,
  and S{\o}gaard}]{ruder2018discriminative}
Sebastian Ruder, Ryan Cotterell, Yova Kementchedjhieva, and Anders S{\o}gaard.
  2018{\natexlab{a}}.
\newblock \href {http://aclweb.org/anthology/D18-1042} {A discriminative
  latent-variable model for bilingual lexicon induction}.
\newblock In \emph{Proceedings of EMNLP}, pages 458--468.

\bibitem[{Ruder et~al.(2018{\natexlab{b}})Ruder, S{\o}gaard, and
  Vuli{\'{c}}}]{Ruder2018survey}
Sebastian Ruder, Anders S{\o}gaard, and Ivan Vuli{\'{c}}. 2018{\natexlab{b}}.
\newblock \href {http://arxiv.org/abs/1706.04902} {{A survey of cross-lingual
  embedding models}}.
\newblock \emph{arXiv preprint arXiv:1706.04902}.

\bibitem[{Sch{\"o}nemann(1966)}]{schonemann1966generalized}
Peter~H Sch{\"o}nemann. 1966.
\newblock A generalized solution of the orthogonal {P}rocrustes problem.
\newblock \emph{Psychometrika}, 31(1):1--10.

\bibitem[{Smith et~al.(2017)Smith, Turban, Hamblin, and
  Hammerla}]{smith2017offline}
Samuel~L. Smith, David~H.P. Turban, Steven Hamblin, and Nils~Y. Hammerla. 2017.
\newblock \href {https://arxiv.org/abs/1702.03859} {Offline bilingual word
  vectors, orthogonal transformations and the inverted softmax}.
\newblock In \emph{Proceedings of ICLR}.

\bibitem[{S{\o}gaard et~al.(2015)S{\o}gaard, Agi{\'{c}}, Mart{\'i}nez~Alonso,
  Plank, Bohnet, and Johannsen}]{Sogaard:2015acl}
Anders S{\o}gaard, {\v{Z}}eljko Agi{\'{c}}, H{\'e}ctor Mart{\'i}nez~Alonso,
  Barbara Plank, Bernd Bohnet, and Anders Johannsen. 2015.
\newblock \href {http://aclweb.org/anthology/P15-1165} {Inverted indexing for
  cross-lingual {NLP}}.
\newblock In \emph{Proceedings of ACL}, pages 1713--1722.

\bibitem[{S{\o}gaard et~al.(2018)S{\o}gaard, Ruder, and
  Vuli{\'{c}}}]{sogaard2018on}
Anders S{\o}gaard, Sebastian Ruder, and Ivan Vuli{\'{c}}. 2018.
\newblock \href {http://aclweb.org/anthology/P18-1072} {On the limitations of
  unsupervised bilingual dictionary induction}.
\newblock In \emph{Proceedings of ACL}, pages 778--788.

\bibitem[{Upadhyay et~al.(2016)Upadhyay, Faruqui, Dyer, and
  Roth}]{upadhyay2016acl}
Shyam Upadhyay, Manaal Faruqui, Chris Dyer, and Dan Roth. 2016.
\newblock \href {http://aclweb.org/anthology/P16-1157} {Cross-lingual models of
  word embeddings: {A}n empirical comparison}.
\newblock In \emph{Proceedings of ACL}, pages 1661--1670.

\bibitem[{Vuli{\'{c}} and Korhonen(2016)}]{vulic2016on}
Ivan Vuli{\'{c}} and Anna Korhonen. 2016.
\newblock \href {http://aclweb.org/anthology/P16-1024} {On the role of seed
  lexicons in learning bilingual word embeddings}.
\newblock In \emph{Proceedings of ACL}, pages 247--257.

\bibitem[{Vuli{\'{c}} and Moens(2015)}]{vulic2015sigir}
Ivan Vuli{\'{c}} and Marie-Francine Moens. 2015.
\newblock {Monolingual and Cross-Lingual Information Retrieval Models Based on
  (Bilingual) Word Embeddings}.
\newblock In \emph{Proceedings of the 38th International ACM SIGIR Conference
  on Research and Development in Information Retrieval (SIGIR)}, pages
  363--372.

\bibitem[{Vuli{\'c} and Moens(2016)}]{vulic2016bilingual}
Ivan Vuli{\'c} and Marie-Francine Moens. 2016.
\newblock \href {https://dl.acm.org/citation.cfm?id=3013583} {Bilingual
  distributed word representations from document-aligned comparable data}.
\newblock \emph{Journal of Artificial Intelligence Research}, 55:953--994.

\bibitem[{Wichmann et~al.(2018)Wichmann, M{\"u}ller, Velupillai, Brown, Holman,
  Brown, Sauppe, Belyaev, Urban, Molochieva et~al.}]{Wichmann:2018asjp}
S{\o}ren Wichmann, Andr{\'e} M{\"u}ller, Viveka Velupillai, Cecil~H Brown,
  Eric~W Holman, Pamela Brown, Sebastian Sauppe, Oleg Belyaev, Matthias Urban,
  Zarina Molochieva, et~al. 2018.
\newblock \href {https://asjp.clld.org/} {The asjp database (version 18)}.

\bibitem[{Williams et~al.(2018)Williams, Nangia, and
  Bowman}]{williams2018broad}
Adina Williams, Nikita Nangia, and Samuel~R. Bowman. 2018.
\newblock A broad-coverage challenge corpus for sentence understanding through
  inference.
\newblock In \emph{Proceedings of the 2018 Conference of the North American
  Chapter of the Association for Computational Linguistics: Human Language
  Technologies, Volume 1 (Long Papers)}, volume~1, pages 1112--1122.

\bibitem[{Xing et~al.(2015)Xing, Wang, Liu, and Lin}]{xing2015normalized}
Chao Xing, Dong Wang, Chao Liu, and Yiye Lin. 2015.
\newblock Normalized word embedding and orthogonal transform for bilingual word
  translation.
\newblock In \emph{Proceedings of the 2015 Conference of the North American
  Chapter of the Association for Computational Linguistics: Human Language
  Technologies}, pages 1006--1011.

\bibitem[{Xu et~al.(2018)Xu, Yang, Otani, and Wu}]{xu2018unsupervised}
Ruochen Xu, Yiming Yang, Naoki Otani, and Yuexin Wu. 2018.
\newblock \href {http://aclweb.org/anthology/D18-1268} {Unsupervised
  cross-lingual transfer of word embedding spaces}.
\newblock In \emph{Proceedings of the 2018 Conference on Empirical Methods in
  Natural Language Processing}, pages 2465--2474. Association for Computational
  Linguistics.

\bibitem[{Yeh(2000)}]{yeh2000more}
Alexander Yeh. 2000.
\newblock More accurate tests for the statistical significance of result
  differences.
\newblock In \emph{Proceedings of the 18th conference on Computational
  linguistics-Volume 2}, pages 947--953. Association for Computational
  Linguistics.

\bibitem[{Zhang et~al.(2017)Zhang, Liu, Luan, and Sun}]{zhang2017adversarial}
Meng Zhang, Yang Liu, Huanbo Luan, and Maosong Sun. 2017.
\newblock Adversarial training for unsupervised bilingual lexicon induction.
\newblock In \emph{Proceedings of ACL}, volume~1, pages 1959--1970.

\bibitem[{Zhang et~al.(2016)Zhang, Gaddy, Barzilay, and
  Jaakkola}]{zhang2016ten}
Yuan Zhang, David Gaddy, Regina Barzilay, and Tommi Jaakkola. 2016.
\newblock \href {http://aclweb.org/anthology/N16-1156} {Ten pairs to tag --
  {Multilingual POS} tagging via coarse mapping between embeddings}.
\newblock In \emph{Proceedings of the 2016 Conference of the North American
  Chapter of the Association for Computational Linguistics: Human Language
  Technologies}, pages 1307--1317.

\bibitem[{Zou et~al.(2013)Zou, Socher, Cer, and Manning}]{zou2013bilingual}
Will~Y Zou, Richard Socher, Daniel Cer, and Christopher~D Manning. 2013.
\newblock Bilingual word embeddings for phrase-based machine translation.
\newblock In \emph{Proceedings of the 2013 Conference on Empirical Methods in
  Natural Language Processing}, pages 1393--1398.

\end{thebibliography}
\bibliographystyle{acl_natbib}

%\clearpage
\setlength{\tabcolsep}{2.2pt}
\begin{table*}[!t]
\centering
\def\arraystretch{0.8}
{\footnotesize
\begin{tabularx}{\linewidth}{l l cc cc cc cc cc cc cc}
\toprule 
Model & Dict & \de-\fin & \de-\fr & \de-\hr & \de-\ita & \de-\ru & \de-\tr & \en-\de & \en-\fin & \en-\fr & \en-\hr & \en-\ita & \en-\ru & \en-\tr & \fin-\fr  \\ \midrule
\cca & 1K & 0.241 & 0.422 & 0.206 & 0.414 & 0.308 & 0.153 & 0.458 & 0.259 & 0.582 & 0.218 & 0.538 & 0.336 & 0.218 & 0.230 \\
\cca & 3K & 0.328 & 0.494 & 0.298 & 0.491 & 0.399 & 0.251 & 0.531 & 0.351 & 0.642 & 0.299 & 0.613 & 0.434 & 0.314 & 0.332 \\
\cca & 5K & 0.353 & 0.509 & 0.318 & 0.506 & 0.411 & 0.280 & 0.542 & 0.383 & 0.652 & 0.325 & 0.624 & 0.454 & 0.327 & 0.362 \\ \midrule
\proc & 1K & 0.264 & 0.428 & 0.225 & 0.421 & 0.323 & 0.169 & 0.458 & 0.271 & 0.579 & 0.225 & 0.535 & 0.352 & 0.225 & 0.239  \\
\proc & 3K & 0.340 & 0.499 & 0.308 & 0.495 & 0.413 & 0.260 & 0.532 & 0.365 & 0.642 & 0.307 & 0.611 & 0.449 & 0.320 & 0.333 \\
\proc & 5K & 0.359 & 0.511 & 0.329 & 0.510 & 0.425 & 0.284 & 0.544 & 0.396 & 0.654 & 0.336 & 0.625 & 0.464 & 0.335 & 0.362 \\ \midrule
\procb & 1K & 0.354 & 0.511 & 0.306 & 0.507 & 0.392 & 0.250 & 0.521 & 0.360 & 0.633 & 0.296 & 0.605 & 0.419 & 0.301 & 0.329 \\
\procb & 3K & 0.362 & 0.514 & 0.324 & 0.508 & 0.413 & 0.278 & 0.532 & 0.380 & 0.642 & 0.336 & 0.612 & 0.449 & 0.328 & 0.350 \\ \midrule
\dlv & 1K & 0.259 & 0.384 & 0.222 & 0.420 & 0.325 & 0.167 & 0.454 & 0.271 & 0.546 & 0.225 & 0.537 & 0.353 & 0.221 & 0.209 \\
\dlv & 3K & 0.341 & 0.496 & 0.306 & 0.494 & 0.411 & 0.261 & 0.533 & 0.365 & 0.636 & 0.307 & 0.611 & 0.444 & 0.320 & 0.321\\
\dlv & 5K &  0.357 & 0.506 & 0.328 & 0.510 & 0.423 & 0.284 & 0.545 & 0.396 & 0.649 & 0.334 & 0.625 & 0.467 & 0.335 & 0.351 \\
\rcsls & 1K & 0.288 & 0.459 & 0.262 & 0.453 & 0.361 & 0.201 & 0.501 & 0.306 & 0.612 & 0.267 & 0.565 & 0.401 & 0.275 & 0.269 \\
\rcsls & 3K & 0.373 & 0.524 & 0.337 & 0.518 & 0.442 & 0.296 & 0.568 & 0.404 & 0.665 & 0.357 & 0.637 & 0.491 & 0.364 & 0.367 \\
\rcsls & 5K & 0.395 & 0.536 & 0.359 & 0.529 & 0.458 & 0.324 & 0.580 & 0.438 & 0.675 & 0.375 & 0.652 & 0.510 & 0.386 & 0.395 \\ \midrule \midrule
\vecmap & -- & 0.302 & 0.505 & 0.300 & 0.493 & 0.322 & 0.253 & 0.521 & 0.292 & 0.626 & 0.268 & 0.600 & 0.323 & 0.288 & 0.368  \\
\muse & -- & 0.000 & 0.005 & 0.245 & 0.496 & 0.272 & 0.237 & 0.520 & 0.000 & 0.632 & 0.000 & 0.608 & 0.000 & 0.294 & 0.348 \\
\icp & -- & 0.251 & 0.454 & 0.240 & 0.447 & 0.245 & 0.215 & 0.486 & 0.262 & 0.613 & 0.000 & 0.577 & 0.259 & 0.000 & 0.000 \\
\gwa & -- & 0.216 & 0.433 & 0.015 & 0.440 & 0.222 & 0.101 & 0.029 & 0.022 & 0.462 & 0.003 & 0.017 & 0.002 & 0.003 & 0.121 \\
\bottomrule
\end{tabularx}
}
\vspace{-2mm}
\caption{\textbf{Appendix:} BLI performance (MAP) for first batch (14) of language pairs.}
\vspace{-5mm}
\end{table*}

%%%
%%%%
%%%

\setlength{\tabcolsep}{2.5pt}
\begin{table*}[!ht]
\centering
\def\arraystretch{0.8}
{\footnotesize
\begin{tabularx}{\linewidth}{l l cc cc cc cc cc cc cc}
\toprule 
Model & Dict & \fin-\hr & \fin-\ita & \fin-\ru & \hr-\fr & \hr-\ita & \hr-\ru & \ita-\fr & \ru-\fr & \ru-\ita & \tr-\fin & \tr-\fr & \tr-\hr & \tr-\ita & \tr-\ru \\ \midrule
\cca & 1K & 0.167 & 0.232 & 0.214 & 0.238 & 0.240 & 0.256 & 0.612 & 0.344 & 0.352 & 0.151 & 0.213 & 0.134 & 0.202 & 0.146 \\
\cca & 3K & 0.264 & 0.328 & 0.306 & 0.346 & 0.345 & 0.348 & 0.659 & 0.452 & 0.449 & 0.232 & 0.308 & 0.211 & 0.309 & 0.252 \\
\cca & 5K & 0.288 & 0.353 & 0.340 & 0.372 & 0.366 & 0.367 & 0.668 & 0.469 & 0.474 & 0.260 & 0.337 & 0.250 & 0.331 & 0.285 \\ \midrule
\proc & 1K & 0.187 & 0.247 & 0.233 & 0.248 & 0.247 & 0.269 & 0.615 & 0.352 & 0.360 & 0.169 & 0.215 & 0.148 & 0.211 & 0.168 \\
\proc & 3K & 0.269 & 0.328 & 0.310 & 0.346 & 0.350 & 0.353 & 0.659 & 0.455 & 0.455 & 0.241 & 0.312 & 0.219 & 0.312 & 0.262 \\
\proc & 5K & 0.294 & 0.355 & 0.342 & 0.374 & 0.364 & 0.372 & 0.669 & 0.470 & 0.474 & 0.269 & 0.338 & 0.259 & 0.335 & 0.290 \\ \midrule
\procb & 1K & 0.263 & 0.328 & 0.315 & 0.335 & 0.343 & 0.348 & 0.665 & 0.467 & 0.466 & 0.247 & 0.305 & 0.210 & 0.298 & 0.230 \\
\procb & 3K & 0.293 & 0.348 & 0.327 & 0.365 & 0.368 & 0.365 & 0.664 & 0.478 & 0.476 & 0.270 & 0.333 & 0.244 & 0.330 & 0.262 \\ \midrule
\dlv & 1K & 0.184 & 0.244 & 0.225 & 0.214 & 0.245 & 0.264 & 0.585 & 0.320 & 0.358 & 0.161 & 0.194 & 0.144 & 0.209 & 0.161 \\
\dlv & 3K & 0.269 & 0.331 & 0.307 & 0.331 & 0.348 & 0.353 & 0.653 & 0.446 & 0.452 & 0.243 & 0.306 & 0.219 & 0.311 & 0.261\\
\dlv & 5K & 0.294 & 0.356 & 0.342 & 0.364 & 0.366 & 0.374 & 0.665 & 0.466 & 0.475 & 0.268 & 0.333 & 0.255 & 0.336 & 0.289 \\ \midrule
\rcsls & 1K & 0.214 & 0.272 & 0.257 & 0.281 & 0.275 & 0.291 & 0.637 & 0.381 & 0.383 & 0.194 & 0.247 & 0.170 & 0.246 & 0.191 \\
\rcsls & 3K & 0.296 & 0.362 & 0.341 & 0.384 & 0.382 & 0.379 & 0.673 & 0.477 & 0.472 & 0.272 & 0.348 & 0.256 & 0.340 & 0.290 \\
\rcsls & 5K & 0.321 & 0.388 & 0.376 & 0.412 & 0.399 & 0.404 & 0.682 & 0.494 & 0.491 & 0.300 & 0.375 & 0.285 & 0.368 & 0.324 \\ \midrule \midrule
\vecmap & -- & 0.280 & 0.355 & 0.312 & 0.402 & 0.389 & 0.376 & 0.667 & 0.463 & 0.463 & 0.246 & 0.341 & 0.223 & 0.332 & 0.200  \\
\muse & -- & 0.228 & 0.000 & 0.001 & 0.000 & 0.000 & 0.000 & 0.662 & 0.005 & 0.450 & 0.000 & 0.000 & 0.133 & 0.000 & 0.000  \\
\icp & -- & 0.208 & 0.263 & 0.231 & 0.282 & 0.045 & 0.309 & 0.629 & 0.000 & 0.394 & 0.173 & 0.000 & 0.138 & 0.243 & 0.119 \\
\gwa & -- & 0.009 & 0.173 & 0.086 & 0.018 & 0.021 & 0.001 & 0.655 & 0.188 & 0.190 & 0.102 & 0.106 & 0.016 & 0.142 & 0.040\\
\bottomrule
\end{tabularx}
}
\vspace{-2mm}
\caption{\textbf{Appendix:} BLI performance (MAP) for the second batch (14) of language pairs.}
\vspace{-2.5mm}
\end{table*}

%\section*{Appendix}

\end{document}